\documentclass[11pt]{article}

\usepackage[margin=1in]{geometry}

\usepackage{amsmath, amssymb, amsthm} 
\usepackage{bm} 
\usepackage{mathtools} 
\usepackage{amsfonts}

\usepackage{graphicx} 
\usepackage{subfig} 
\usepackage{tabularx}
\usepackage{booktabs}
\usepackage{ragged2e}

\usepackage[auth-lg]{authblk}

\usepackage{hyperref}
\hypersetup{
    colorlinks=true,
    linkcolor=blue,
    urlcolor=cyan,
    citecolor=blue,
    pdftitle={Explaining Deep Network Classification of Matrices: A Case Study on Monotonicity},
    pdfauthor={Leandro Farina, Sergey Korotov},
}

\theoremstyle{remark}
\newtheorem{remark}{Remark}[section] 

\title{Explaining Deep Network Classification of Matrices: A Case Study on Monotonicity}

\author[1]{Leandro Farina\thanks{\texttt{leandro.farina@gmail.com} (Corresponding author)}}
\author[2]{Sergey Korotov\thanks{\texttt{sergey.korotov@mdu.se}}}

\affil[1]{Institute of Mathematics and Statistics, Federal University of Rio Grande do Sul, Porto Alegre, RS, Brazil}
\affil[2]{Division of Mathematics and Physics, UKK, Mälardalen University, Västerås, Sweden}

\date{}

\begin{document}

\maketitle

\begin{abstract}
This work demonstrates a methodology for using deep learning to discover simple, practical criteria for classifying matrices based on abstract algebraic properties. By combining a high-performance neural network with explainable AI (XAI) techniques, we can distill a model's learned strategy into human-interpretable rules. We apply this approach to the challenging case of monotone matrices, defined by the condition that their inverses are entrywise nonnegative. Despite their simple definition, an easy characterization in terms of the matrix elements or the derived parameters is not known. Here, we present, to the best of our knowledge, the first systematic machine-learning approach for deriving a practical criterion that distinguishes monotone from non-monotone matrices. After establishing a labelled dataset by randomly generated monotone and non-monotone matrices uniformly on $(-1,1)$, we employ deep neural network algorithms for classifying the matrices as monotone or non-monotone, using both their entries and a comprehensive set of matrix features. By saliency methods, such as integrated gradients, we identify among all features, two matrix parameters which alone provide sufficient information for the matrix classification, with $95\%$ accuracy, namely the absolute values of the two lowest-order coefficients, $c_0$ and $c_1$ of the matrix's characteristic polynomial. A data-driven study of 18,000 random $7\times7$ matrices shows that the monotone class obeys $\lvert c_{0}/c_{1}\rvert\le0.18$ with probability $>\!99.98\%$; because $\lvert c_{0}/c_{1}\rvert = 1/\mathrm{tr}(A^{-1})$ for monotone $A$, this is equivalent to the simple bound $\mathrm{tr}(A^{-1})\ge5.7$.

\vspace{1em}
\noindent\textbf{Keywords:}
  Monotone matrices; neural networks; explainable AI; saliency; extreme-value theory; symbolic regression.

\noindent\textbf{MSC Codes:}
  15B48, 68T07, 15A18, 62G32
\end{abstract}


\section{Introduction}
Although there are different definitions of a monotone matrix, here we are concerned with the following:  a real square matrix $A$ is said to be monotone if $A \bm{x}_2 \geq A \bm{x}_1$ implies $\bm{x}_2 \geq  \bm{x}_1$, for any two vectors $\bm{x}_1$ and $\bm{x}_2$ \cite{collatz}.  The inequalities here are valid element-wise.  This definition is equivalent to the statement that  $A$ is monotone if and only if  $A^{-1} \geq 0$, see e.g. \cite{mangasarian}.  For this reason, monotone matrices are sometimes referred to as inverse nonnegative matrices.  Analogous properties are found in continuous problems, for instance for the Laplace operator, the following result (called the maximum principle) holds: if \(v(x,y)\) is a sufficiently smooth function on a region \(R\) with boundary \(C\), and  
\[
-\Delta v \ge 0\quad\text{in }R,
\qquad
v\ge 0\quad\text{on }C,
\]  
then  
\[
v\ge 0\quad\text{throughout }R.
\]  
This is why the concept of monotone matrices is widely employed in various works devoted to the so-called discrete maximum principles (discrete analogs of maximum principles) for the finite element and finite difference approximations of elliptic and parabolic equations (and systems of such equations), see e.g. \cite{ciarlet, ciarlet-raviart, karkor} and references therein.

Unfortunately, there is no characterization of the whole class of monotone matrices. Moreover, in practice, in the context of  finite element and difference discretizations, one often addresses to  some special sub-classes of monotone matrices, such as M-matrices or Stieltjes matrices, see \cite[p.~91]{varga} and \cite{brakorkri}. In these classes, the off-diagonal entries are required to be non-positive, which leads to various geometric conditions on the meshing of domains (cf.~\cite{karkor}). Several sets of (sufficient) conditions on entries and properties of the matrix guaranteeing its monotonicity, and generalizing the concepts of M-matrix (especially avoiding the requirement of non-positivity of off-diagonal entries), have been found and presented e.g. in the works \cite{bouchon, bramble-hubbard, lorenz}.

A particular and well-studied type of monotone matrix is called an M‑matrix and is defined by the property that it is a Z‑matrix (i.e.\ all its off‑diagonal entries are non‑positive) whose inverse is entry‑wise non‑negative.  Equivalently, \(A\) is an M‑matrix if and only if it can be written in the form
\[
A \;=\; sI \;-\; B,
\]
where \(B\ge0\) and \(s\ge\rho(B)\) (the spectral radius of \(B\)), which guarantees \(A^{-1}\ge0\).

Artificial intelligence (AI) is increasingly utilized in mathematical research, with interpretability methods like saliency maps offering paths to human-understandable insights and even new conjectures, as demonstrated in fields like knot theory \cite{sundararajan2017axiomatic, davies2021advancing}. The expanding role of AI, from advanced symbolic manipulations \cite{novikov2025alphaevolve} to tools for mathematical reasoning and conjecture generation \cite{williamson2023deep, dwivedi2023explainable, ghrist2024lattice, buzzard2024mathematical}, underscores its potential. However, AI models can also master complex tasks without necessarily providing explanations for their success \cite{charton2022linear}. 
The present work contrasts with the latter by employing AI not only to classify monotone matrices but, crucially, to identify the minimal, interpretable structural features underpinning this classification, thereby seeking to {\em explain} the learned mathematical property.

 Section \ref{sec:dataset} formalises the binary classification task and details the construction of balanced training sets for $n=5,7,10$.  Section \ref{sec:singlefeature} benchmarks baseline neural architectures that use only the raw matrix entries, while Section \ref{sec:derivedfeat} augments those inputs with a set of algebraic and spectral descriptors, forming a high-accuracy 73-feature model.  In Section \ref{sec:salience_and_features} we aim to turn the network inside-out with integrated gradients, decision-tree surrogates, and symbolic regression, revealing that just two characteristic-polynomial coefficients, $|c_0|$ and $|c_1|$, carry almost the entire monotonicity signal.  
 Sections \ref{sec:twofeat}–\ref{sec:reduction} test this insight directly: stripped-down two- and one-feature networks match the best models’ recall while generalising even better, and extreme-value analysis of their single-feature ratio yields a sharp, data-driven necessary condition for monotonicity in the $7\times7$ case.  Section \ref{sec:separation_search} compares alternative linear and non-linear separators, and Section \ref{sec:filtered_subdomain} shows how a hierarchical approach refines classification inside the low-ratio subdomain where the coarse test is already favourable.  Finally, Section \ref{sec:conclusion} summarizes the main findings and charts future work, including targeted datasets for structured subclasses such as M- and Stieltjes matrices.
\section{The Problem and the Dataset}\label{sec:dataset}
We consider the binary classification problem of deciding if a matrix \(A\in\mathbb{R}^{n\times n}\), with entries drawn independently from \(\mathcal{U}(-1,1)\), is \emph{monotone}, i.e., \(A^{-1}\) exists and $(A^{-1})_{ij}\ge0 \: \forall\,i,j$. Classifier performance is assessed using standard metrics (Accuracy, Precision, Recall, Specificity) derived from the confusion matrix (TP, TN, FP, FN counts).

We generated labeled data for \(n=5, 7,\) and \(10\). Matrices $A \sim \mathcal{U}(-1,1)^{n\times n}$ were drawn, their monotonicity was determined by checking if $\det(A)\neq0 \wedge A^{-1}_{ij}\ge0 \: \forall\,i,j$, and rejection sampling was used to construct balanced datasets. This yielded $18\,000$ matrices each for $n=5, 7$, and $2\,500$ for $n=10$ (reflecting increasing rarity, discussed below), split evenly between monotone and non-monotone classes. Each accepted matrix was stored in a compressed HDF5 dataset using NumPy/CuPy.

Monotone matrices are exponentially rare as $n$ increases. Approximating the probability that all $n^2$ entries of $A^{-1}$ (from a random invertible $A$) are non-negative (assuming entries are roughly symmetric about zero and signs are independent) gives $\mathbb{P}(A \text{ monotone}) \approx (\frac{1}{2})^{n^2}$. For $n=10$, this is $\approx 7.9 \times 10^{-31}$, explaining our limitation to $n=10$ given available compute.
\section{Classification by Deep Neural Networks}\label{sec:dnn}

In this section we describe our deep neural network approach to classifying matrices as monotone or non‐monotone.  All experiments in what follows use the matrices of sizes 5, 7 and 10, focusing on the $7 \times 7$ case, splitting 80/20 into training and validation sets. 
%
\subsection{Only the raw matrix entries as features} \label{sec:singlefeature}

We first considered classifying matrices using only their raw entries (flattened $7 \times 7 = 49$ features). We compared a feed-forward network (FFN), which treats entries as a generic feature vector, with a convolutional neural network (CNN) designed to exploit potential spatial correlations. Architectures are detailed in Table~\ref{tab:dnn-architectures-split}.

Both models used binary cross-entropy loss, the Adam optimizer ($10^{-3}$ learning rate), batch size 32, and trained for up to 25 epochs with early stopping (validation loss patience 6), using TensorFlow’s \texttt{tf.data} pipeline.

\begin{figure}[htbp]
%
  \centering
  \includegraphics[width=0.9\textwidth]{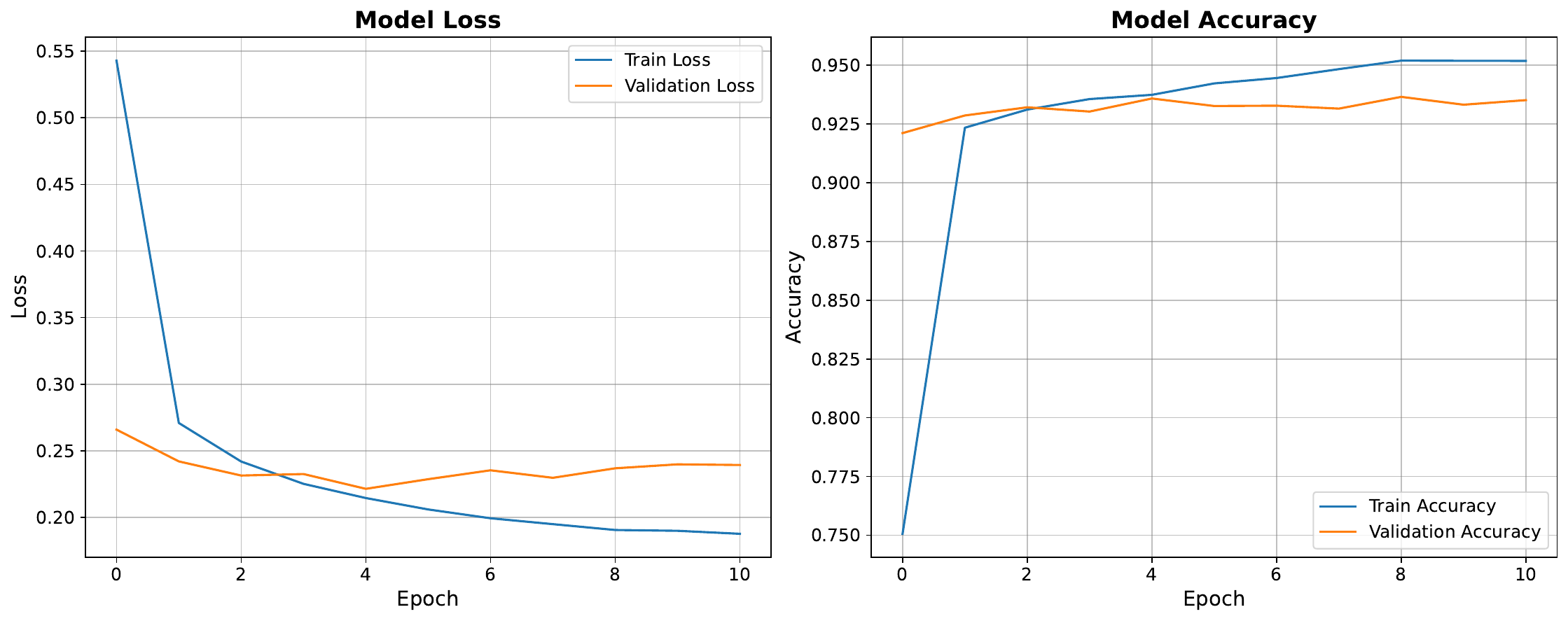} 
  \caption{Evolution of training and validation loss and accuracy over epochs for the 7$\times$7 FFN matrix classifier using raw entries.}
  \label{fig:perf-7x7}
\end{figure}

Figure~\ref{fig:perf-7x7} shows the FFN model training evolution; by epoch~8, it achieved over 93\% validation accuracy from raw entries. Some overfitting was apparent, addressed later with derived features (Sections~\ref{sec:derivedfeat}, \ref{sec:twofeat}).

\begin{table}[htbp]
  \centering
  \caption{Architectures of FFN and CNN classifiers for the 7$\times$7 dataset using raw matrix entries. Both use ReLU activation unless noted and share the same final head structure.}
  \label{tab:dnn-architectures-split}
  \begin{tabularx}{\textwidth}{@{} l l >{\RaggedRight}X >{\RaggedRight}X l @{}}
    \toprule
    Model & Input & Core Architecture & Regularization & Final Head \\
    \midrule
    FFN & Flattened (49)
        & Dense(1024) $\to$ Dense(512) $\to$ Dense(256) $\to$ Dense(128) $\to$ Dense(64) $\to$ Dense(32)
        & L2 ($\lambda=10^{-4}$), Dropout(0.25) on hidden layers
        & \textit{Shared} \\
    \addlinespace
    CNN & Reshaped (7,7,1)
        & Conv2D(32, 3x3, same) $\to$ MaxPool(2x2) $\to$ Conv2D(64, 3x3, same) $\to$ MaxPool(2x2) $\to$ Flatten $\to$ Dense(128)
        & L2 on Dense(128), Dropout(0.2) on Dense(128)
        & \textit{Shared} \\
    \bottomrule
  \end{tabularx}
  \par
  \footnotesize{\textit{Shared Head}: The output of the model-specific layers (Dense(32) for FFN, Dense(128) for CNN) is passed to a final common sequence: Dense(32) $\to$ Dropout $\to$ Sigmoid output neuron.}
\end{table}

On the full dataset ($36,000$ matrices), the FFN (confusion matrix: $\begin{bsmallmatrix} 17297 & 703\\ 412 & 17588 \end{bsmallmatrix}$) outperformed the CNN ($\begin{bsmallmatrix} 16613 & 1387\\ 1065 & 16935 \end{bsmallmatrix}$); see Table~\ref{tab:dnn-metrics-split} for performance metrics. This suggested that simple spatial correlations captured by our CNN were not decisively informative for monotonicity here. Further experiments with more sophisticated CNNs (e.g., with Squeeze-and-Excitation blocks, achieving $\approx 94.4\%$ accuracy) still did not match the FFN's performance. Consequently, we used the simpler FFN as our reference model for analysis with raw entries.

\begin{table}[htbp]
%
%
  \centering
  \caption{Performance comparison of FFN and CNN classifiers on the 7$\times$7 dataset using raw matrix entries. Best performance metrics are in bold.}
  \label{tab:dnn-metrics-split}
  \begin{tabular}{@{}lcccc@{}} 
    \toprule
    Model & Accuracy & Precision & Recall & Specificity \\
    \midrule
    FFN   & \textbf{96.9\%} & \textbf{96.2\%} & \textbf{97.7\%} & \textbf{96.1\%} \\
    CNN   & 93.2\%          & 92.4\%          & 94.1\%          & 92.3\%          \\
    \bottomrule
  \end{tabular}
\end{table}

\subsection{Including derived features} \label{sec:derivedfeat}
To potentially improve classification, we considered derived features that summarize algebraic or spectral matrix properties. Our initial comprehensive set of $N_{\text{derived}} = 5 + 4n$ potential features for an $n \times n$ matrix $A$ is detailed in Table~\ref{tab:derived-features-full}. This set includes quantities like spectral radius, trace, Fiedler value, eigenvalues (real/imaginary parts), characteristic polynomial coefficients ($c_k$), their absolute values ($|c_k|$), and ratios of low-order coefficients. Each feature type is assigned an identifier (ID) for reference. 

\begin{table}[htbp]
  \centering
  \caption{Derived features initially computed and considered, derived from \(A \in \mathbb{R}^{n\times n}\).}
  \label{tab:derived-features-full}
  \begin{tabularx}{\textwidth}{@{} l l >{\RaggedRight}X c @{}}
    \toprule
    ID & Feature Name & Description / Notation & Count \\
    \midrule
    DF1 & Spectral Radius & Largest absolute eigenvalue: \(\rho(A) = \max_i |\lambda_i(A)|\) & 1 \\
    \addlinespace
    DF2 & Fiedler Value & Second smallest eigenvalue of the associated graph Laplacian (\(\lambda_2(L)\)), related to algebraic connectivity. & 1 \\
    \addlinespace
    DF3 & Trace & Sum of diagonal elements: \(\mathrm{tr}(A) = \sum_{i=1}^n A_{ii}\) & 1 \\
    \addlinespace
    DF4 & Eigenvalues (Real Parts) & Real parts of the \(n\) eigenvalues: \(\mathrm{Re}(\lambda_i(A))\) & \(n\) \\
    \addlinespace
    DF5 & Eigenvalues (Imaginary Parts) & Imaginary parts of the \(n\) eigenvalues: \(\mathrm{Im}(\lambda_i(A))\) & \(n\) \\
    \addlinespace
    DF6 & CharPoly Coefficients & Coefficients \(c_k\), \(k=0,..., n-1\) & \(n\) \\
    \addlinespace
    DF7 & Abs(CharPoly Coefficients) & Absolute values: \(|c_k|\), \(k=0,...,n-1\) & \(n\) \\
    \addlinespace
    DF8 & Coefficient Ratio 1 & Ratio: \( |c_0 / c_1| \) (handled for \(c_1=0\)) & 1 \\
    \addlinespace
    DF9 & Coefficient Ratio 2 & Ratio: \(|c_0 / (c_1 + c_2)|\) (handled for denominator=0) & 1 \\
    \midrule
    \multicolumn{3}{@{}l}{\textbf{Total Initially Considered Derived Features}} & \textbf{5 + 4n} \\
    \bottomrule
  \end{tabularx}
\end{table}

While the full set listed in Table~\ref{tab:derived-features-full} was available, preliminary analysis led to a refined feature set. Raw characteristic polynomial coefficients (DF6) were excluded due to sign complexity, their magnitudes being captured by absolute values (DF7). Coefficient ratios like $|c_0/c_1|$ (DF8) were set aside for focused analysis in Section~\ref{sec:reduction}, given their overlap with $|c_0|$ and $|c_1|$.
Thus, the core derived feature set combined with raw entries comprises the Spectral Radius (DF1), Fiedler Value (DF2), Trace (DF3), Eigenvalue Real/Imaginary parts (DF4/DF5), and Absolute CharPoly Coefficients (DF7). This selection constitutes $3+3n$ derived features. We next examine the effectiveness of this refined set, before returning to the role of $|c_0|$, $|c_1|$ from DF7 and the DF8 ratio. 

We then tested if combining these selected derived features with raw matrix entries yielded better results. For $n=7$, a hybrid input vector of $3+3n=24$ derived features plus $n^2=49$ raw entries (total 73 features) was created.
An FFN classifier, architecturally similar to that in Table~\ref{tab:dnn-architectures-split} but with an additional 2056-neuron dense layer (before the 1024-neuron layer) and dropout adjusted to 0.35 for the larger input, was trained on this hybrid set. Standard training procedures and validation remained as in Section~\ref{sec:singlefeature}. 

\begin{figure}[htbp]
%
%
  \centering
  \includegraphics[width=0.9\textwidth]{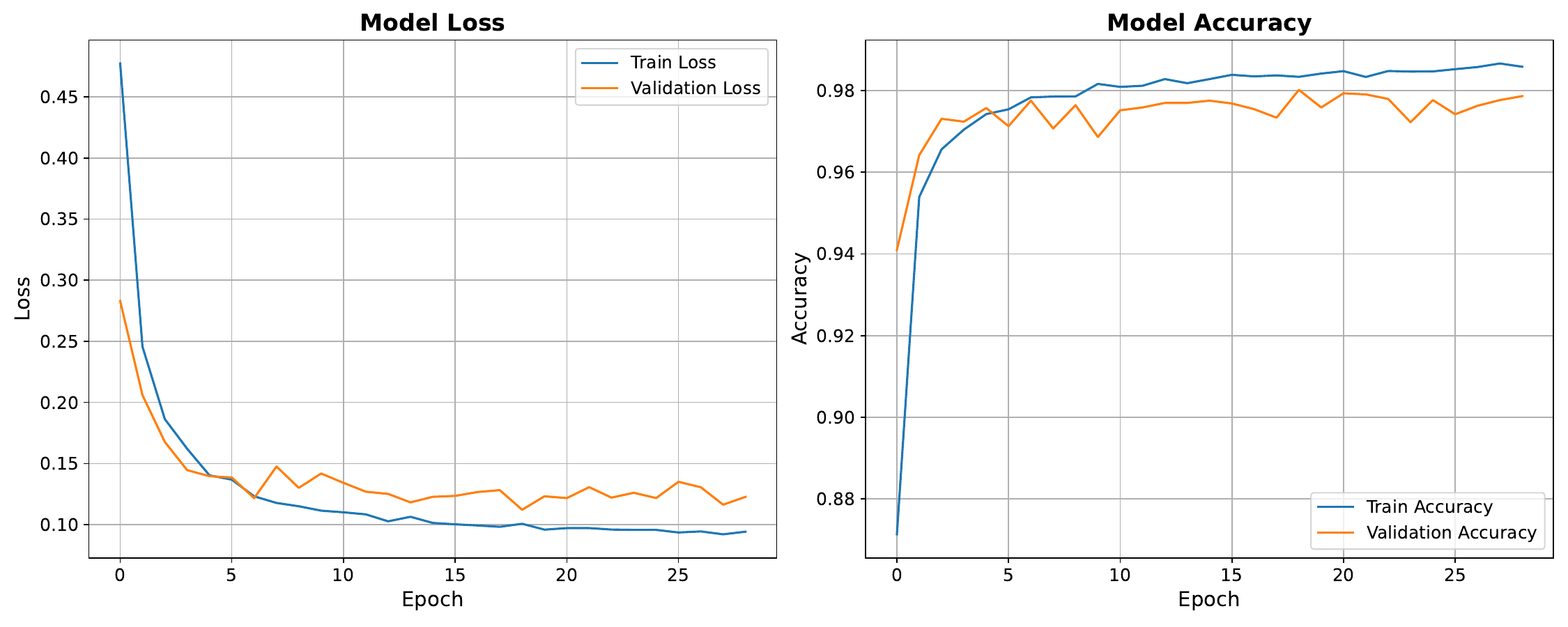}
  \caption{Evolution of training and validation loss and accuracy over epochs for the 7$\times$7 FFN matrix classifier using a hybrid input of selected derived features (DF1-DF5, DF7) and raw matrix entries.}
  \label{fig:perf-derived}
\end{figure}

Figure~\ref{fig:perf-derived} shows this hybrid-input model training. It converged rapidly, and compared to the raw-entry FFN (Fig.~\ref{fig:perf-7x7}), exhibited significantly lower loss and less overfitting. The model yielded the confusion matrix $\begin{bsmallmatrix} 17825 & 175\\ 170 & 17830 \end{bsmallmatrix}$ on the full training set. 
The hybrid model (Table~\ref{tab:dnn-metrics-derived}) substantially improved classification over the raw-entry FFN (Table~\ref{tab:dnn-metrics-split}), with accuracy increasing from 96.9\% to 99.0\% and better generalization. This high-accuracy hybrid FFN became our primary model for extracting interpretable insights, as detailed in subsequent sections.

\begin{table}[htbp]
  \centering
  \caption{Performance of FFN classifier on the 7$\times$7 dataset using a hybrid input of selected derived features (DF1-DF5, DF7) and raw matrix entries. Best performance metrics are in bold.}
  \label{tab:dnn-metrics-derived}
  \begin{tabular}{@{}lcccc@{}}
    \toprule
    Model Input & Accuracy & Precision & Recall & Specificity \\
    \midrule
     Hybrid Input (FFN) & \textbf{99.0\%} & \textbf{99.0\%} & \textbf{99.1\%} & \textbf{99.0\%} \\
    \bottomrule
  \end{tabular}
\end{table}

\section{The two most explanatory features}\label{sec:salience_and_features}
%

\subsection{Salience}

Aiming for human-interpretable understanding, we employed explainable AI (XAI) feature attribution methods, specifically Integrated Gradients (IG)~\cite{sundararajan2017axiomatic}. IG assigns an importance (salience) score to each input feature based on its contribution to the model output, overcoming saturation issues common with simple gradients where highly influential features might be wrongly assigned zero importance.
IG achieves this by integrating gradients of the model output \(F\) with respect to features along a straight-line path from a chosen baseline \(x'\) (e.g., an average input) to the actual input \(x\). The attributed score for feature \(x_i\) is
\[
\text{IG}_i(x, x') = (x_i - x'_i)  \int_{0}^{1} \frac{\partial F(x' + \alpha(x - x'))}{\partial x_i} \, d\alpha.
\]
Numerically approximated by summing gradients at discrete steps, IG notably adheres to {\em Completeness} ($\sum_i \text{IG}_i(x, x') = F(x) - F(x')$), ensuring a full accounting of the output change distributed among input features.

\begin{figure}[htbp]
  \centering
  %
  %
  \includegraphics[width=0.9\textwidth]{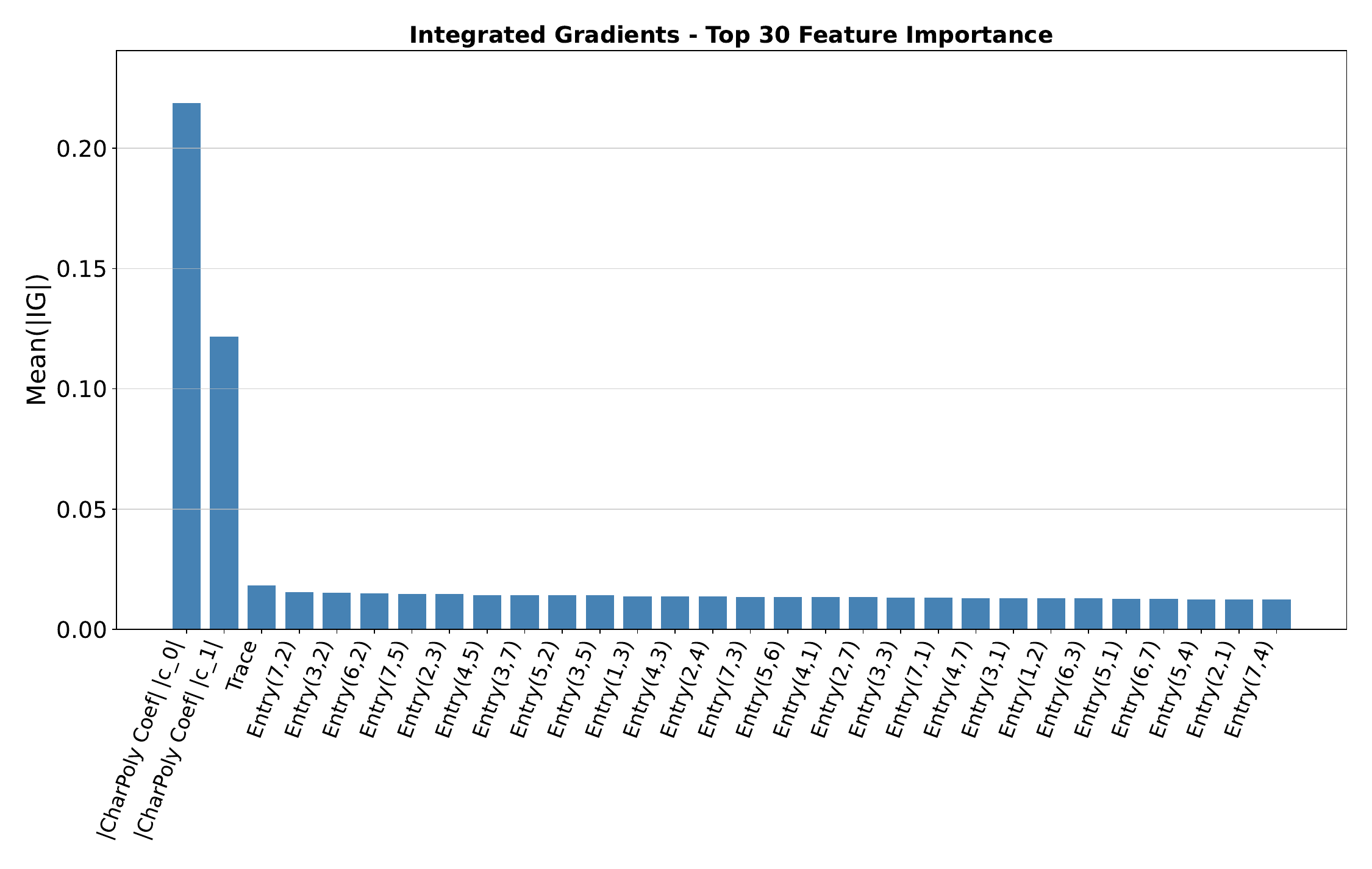}
  \caption{Feature-importance ranking based on the mean absolute Integrated Gradients score (\(|\text{IG}|\)), computed for the hybrid FFN model (7$\times$7 matrices, derived features DF1-DF5, DF7 plus raw entries) across the dataset using 100 steps. The plot shows the top 30 features.}
  \label{fig:ig_summary}
\end{figure}

We computed IGs for the hybrid FFN model (73 features: DF1-DF5, DF7 plus \(n^2\) raw entries) across the entire dataset, using mean feature values as baseline and \(m=100\) integration steps. Figure~\ref{fig:ig_summary} shows the mean absolute IG scores for the top 30 features.

The IG results (Fig.~\ref{fig:ig_summary}) are striking: the absolute values of the two lowest-order characteristic polynomial coefficients, \(|c_0|\) and \(|c_1|\) (from DF7, see Table~\ref{tab:derived-features-full}), overwhelmingly dominate. \(|c_0|\) has the highest mean absolute IG, followed by \(|c_1|\), after which importance drops abruptly. The Trace (DF3) and matrix entry \(A_{7,2}\) are next, with other features (e.g., spectral radius, Fiedler value, other eigenvalue components, or most raw entries) showing substantially lower salience and not appearing in the top 30.
This analysis strongly suggests the hybrid FFN, despite access to 73 features, primarily relies on \(|c_0|\) and \(|c_1|\) for its high classification accuracy.

Recalling the mathematical meaning of these coefficients from $p(\lambda) = \det(\lambda I - A) =  \lambda^n + c_{n-1}\lambda^{n-1} + \dots + c_1\lambda + c_0$: the constant term \(c_0 = (-1)^n \det(A)\), so \(|c_0|=|\det(A)| = \prod_{i=1}^n |\lambda_i|\) by Vieta's formulas. The linear term coefficient, \(c_1\), is related to the sum of the $(n-1) \times (n-1)$ principal minors, equivalent to $c_1 = (-1)^{n-1} \mathrm{tr}(\mathrm{adj}(A))$, where \(\mathrm{adj}(A)\) is the adjugate matrix. In terms of eigenvalues, \(c_1\) also corresponds to \((-1)^{n-1}\) times the sum of the products of all combinations of \(n-1\) eigenvalues. The IG analysis thus indicates the network uses global information encoded in \(|c_0|\) and \(|c_1|\), rather than relying directly on individual eigenvalues, the overall trace $\mathrm{tr}(A)$ (which is $-c_{n-1}$), or specific raw matrix entries beyond $A_{7,2}$. This preference for low-order polynomial coefficients offers insight into the network learned properties for monotonicity.

\begin{figure}[htbp]
  \centering
  \subfloat[\(n=5\)]{\label{fig:scatter_n5}\includegraphics[width=0.325\textwidth]{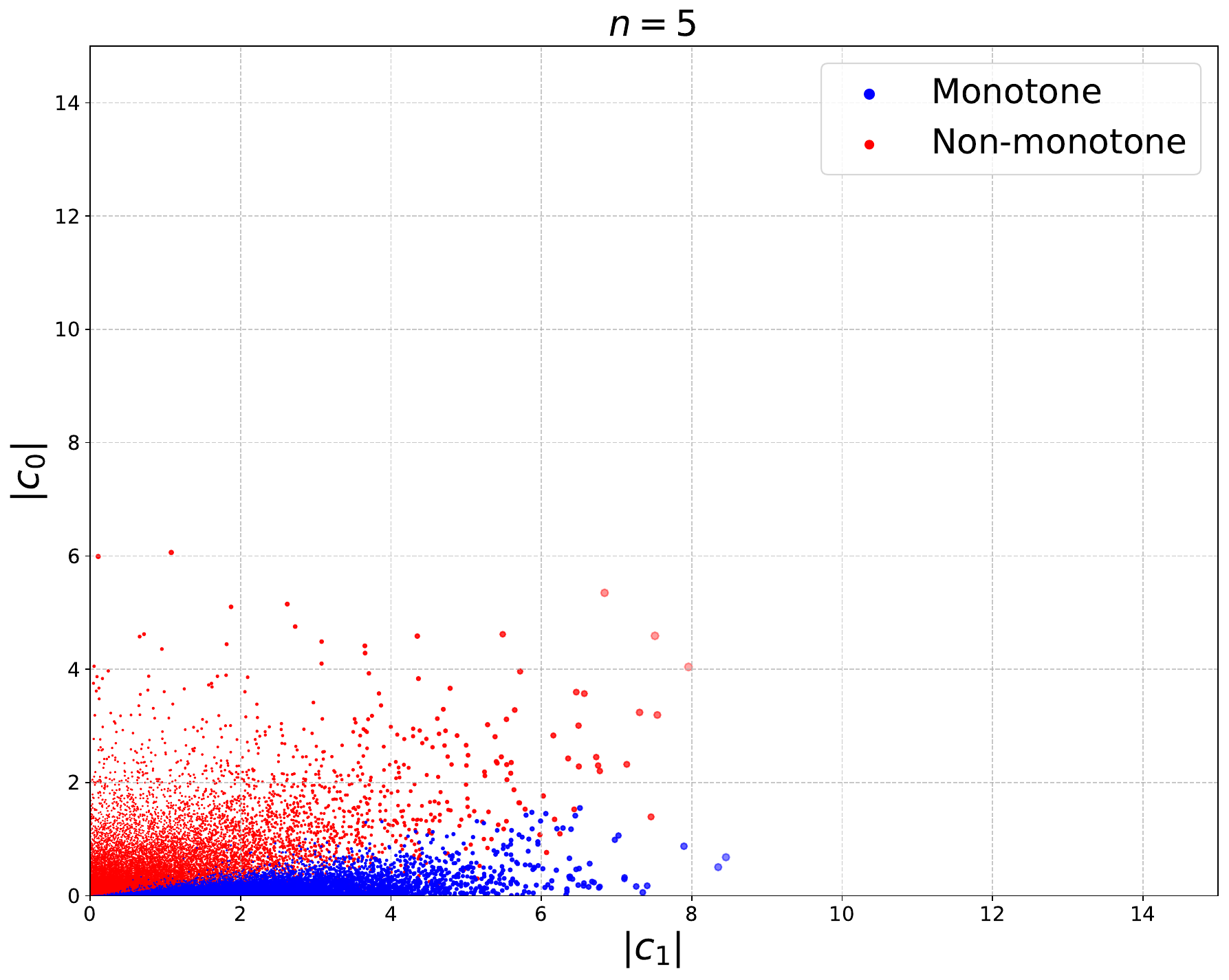}}
  \hfill 
  \subfloat[\(n=7\)]{\label{fig:scatter_n7}\includegraphics[width=0.325\textwidth]{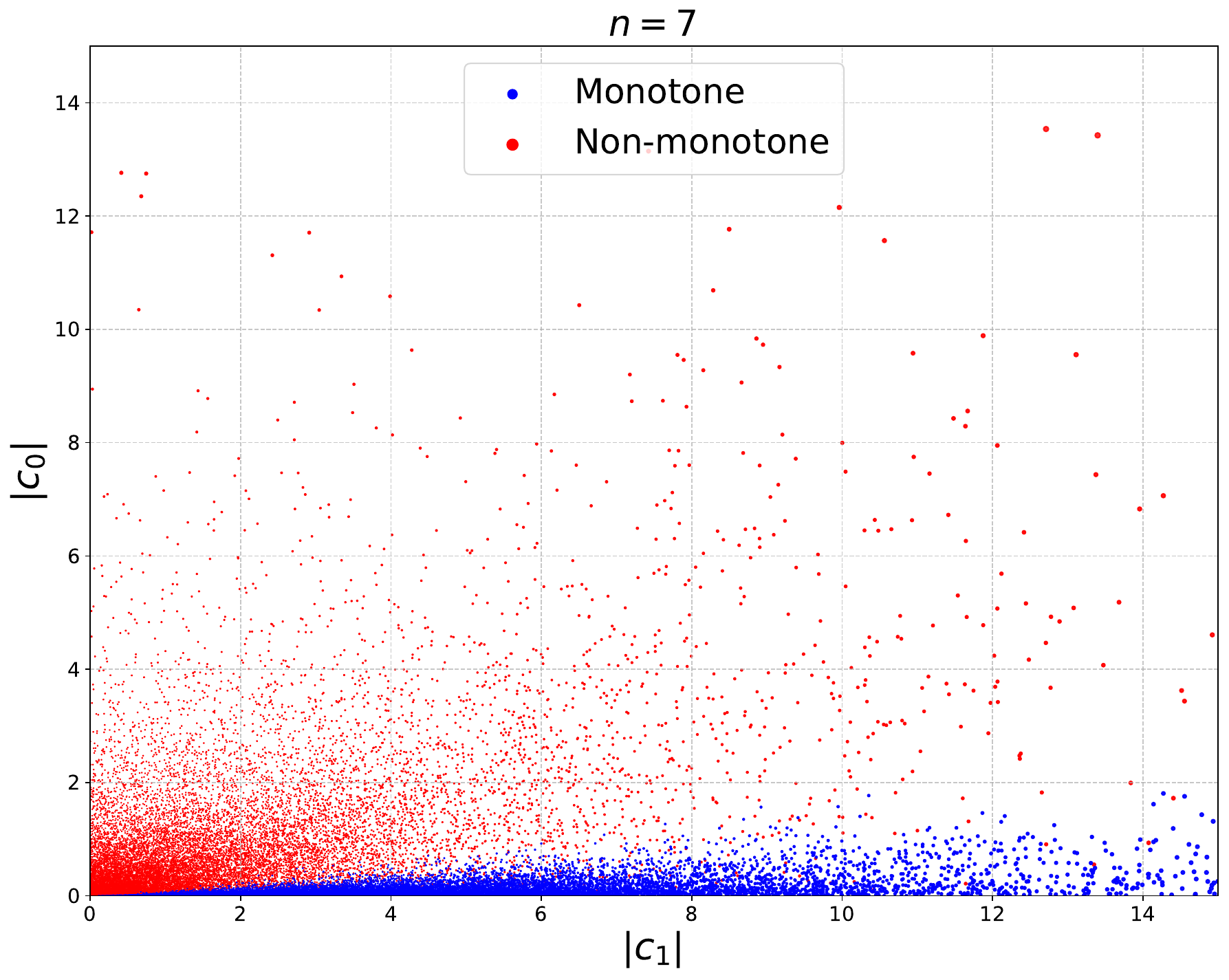}}
  \hfill 
  \subfloat[\(n=10\)]{\label{fig:scatter_n10}\includegraphics[width=0.325\textwidth]{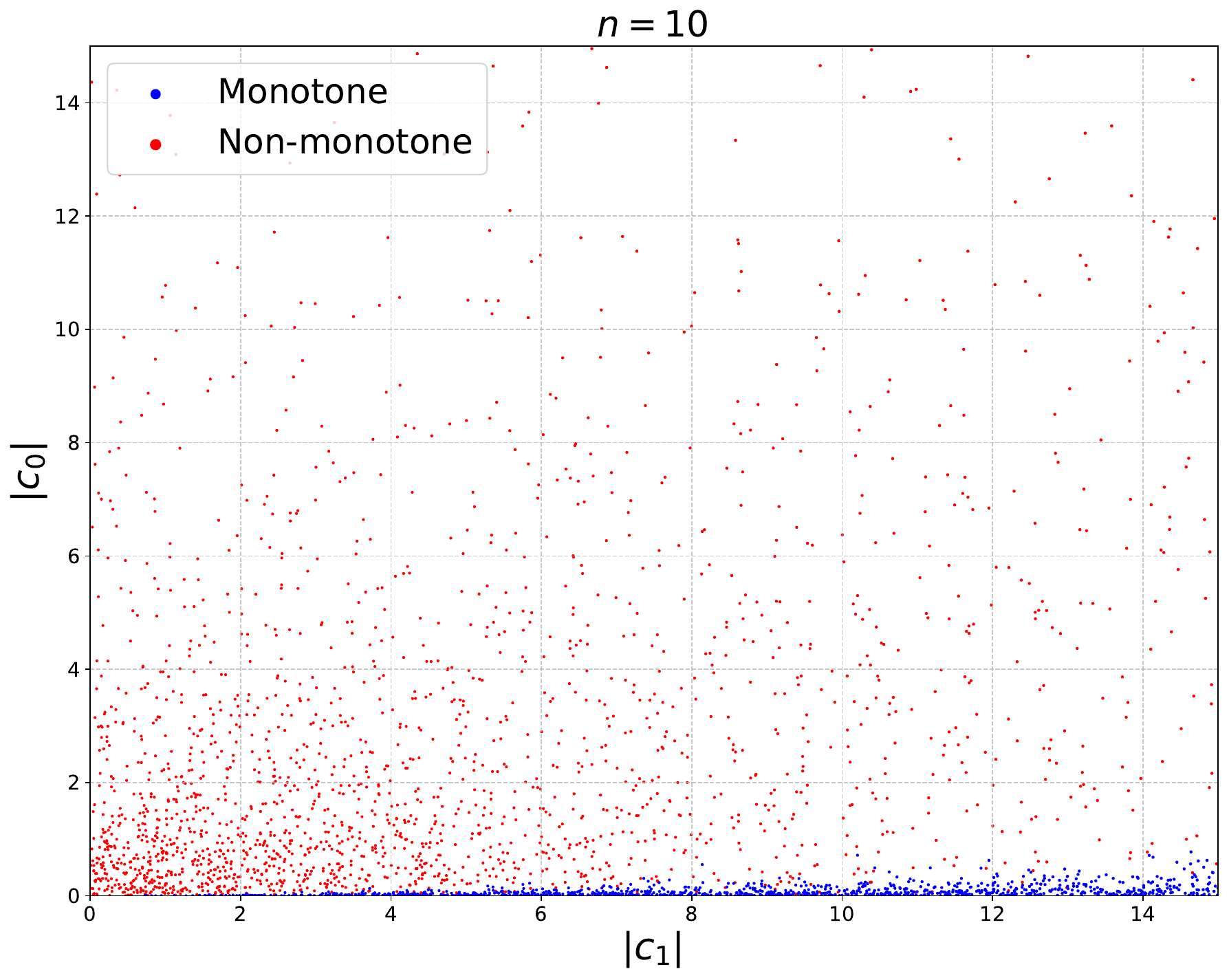}}
  \caption{Scatter plots of \(|c_0|\) versus \(|c_1|\) for monotone (blue) and non-monotone (red) matrices in the datasets for matrix sizes \(n=5, 7,\) and \(10\). Note the concentration of monotone matrices near the origin.}
  \label{fig:scatter_c0_c1}
\end{figure}

To further visualize the importance and relationship of these two dominant features, Figure~\ref{fig:scatter_c0_c1} presents scatter plots of \(|c_0|\) versus \(|c_1|\) for monotone (blue) and non-monotone (red) matrices in our datasets ($n=5, 7, 10$).

%
%

These scatter plots (Fig.~\ref{fig:scatter_c0_c1}) confirm the predictive strength of \(|c_0|\) and \(|c_1|\). Monotone matrices (blue) predominantly occupy a narrow band of very small \(|c_0|\) (i.e., \(|\det(A)|\)) values, though extending over a wide range of \(|c_1|\) values. Non-monotone matrices (red) are more broadly distributed, significantly occupying regions with larger \(|c_0|\). However, the separation is imperfect and non-linear: non-monotone points extend into the low-\(|c_0|\) region, and some monotone points are mixed with non-monotone ones at lower \(|c_1|\) values. As \(n\) increases, the non-monotone point density appears to rise, making the boundary more complex.
The imperfect separation, despite the high importance of these features, suggests that more sophisticated methods or additional, less salient features might be needed for perfect classification, motivating explorations in Sections~\ref{sec:separation_search} and~\ref{sec:filtered_subdomain}.

\subsection{Decision trees}\label{sec:dt}
To obtain an \emph{explicit rule set} that a human can audit, we fitted a
CART--style decision tree \cite{breiman1984cart} to the \emph{predicted}
labels of the hybrid FFN.\footnote{%
  Training on the FFN’s outputs (rather than on the ground-truth labels)
  gives a \emph{surrogate} that mimics the network’s boundary; this is the
  “pedagogical” rule–extraction paradigm.
}
The implementation used
\texttt{sklearn.tree.DecisionTreeClassifier} with
Gini impurity and a maximum depth of~4, deep enough to capture the main
structure while keeping the rule set compact.
The tree was trained on the full \(36\,000\) matrices of size \(7\times7\)
and supplied with all 73 input features
(24 derived + 49 raw).  
Remarkably, every split selected by CART involved only
the two coefficients
\(\,r=|c_{0}|\) (determinant magnitude) and
\(\,s=|c_{1}|\) (related to the trace of the adjugate).

Collecting the leaf nodes labelled monotone yields the empirical
\emph{piece-wise sufficient} condition in Table~\ref{tab:dt-bands-depth4}.
Whenever a matrix satisfies \emph{any} row of the table, the hybrid FFN
(and therefore the surrogate tree) classifies it as monotone.

\begin{table}[htbp]
  \centering
  \caption{Axis‑aligned rectangles in the $(|c_{0}|,|c_{1}|)$ plane that correspond to monotone leaves of the depth‑4 decision tree.  Here $r=|c_{0}|$ and $s=|c_{1}|$.}
  \label{tab:dt-bands-depth4}
  \begin{tabular}{@{}ccc@{}}
    \toprule
    Band & $r$ range & required $s$ range \\ \midrule
    1 & $0 \le r \le 0.02$ & $0.30 < s \le 0.97$ \\
    2 & $0 \le r \le 0.07$ & $0.97 < s \le 1.66$ \\
    3 & $0 \le r \le 0.22$ & $1.66 < s \le 2.72$ \\
    4 & $0.22 < r \le 0.38$ & $3.57 < s \le 4.58$ \\
    5 & $0.22 < r \le 0.93$ & $s > 4.58$ \\ \bottomrule
  \end{tabular}
\end{table}

 The dominance of \(|c_0|\) and \(|c_1|\) in both Integrated Gradients (Fig.~\ref{fig:ig_summary})
and the decision tree supports our earlier conclusion:
for matrices sampled uniformly from \((-1,1)^{7\times7}\), the magnitudes of the lowest-order characteristic polynomial coefficients, \(|c_0|\) and \(|c_1|\), are \emph{sufficient statistics} for the learned monotonicity classifier.
\subsection{Symbolic regression}\label{sec:symreg}
To test whether the surrogate decision surface admits a compact closed‐form description, we applied \emph{symbolic regression} with the open-source package PySR\,\cite{cranmer2020pysr}.  Each matrix was encoded by the full 73-dimensional feature vector used by the hybrid network, and the network binary output \(y_{\text{FFN}}\in\{0,1\}\) was mapped to the discriminant target \(p=2y_{\text{FFN}}-1\in\{-1,+1\}\).  PySR explored algebraic expressions built from the terminals \(x_{0},\dots ,x_{72}\) (where \(x_{22}=|c_{1}|\) and \(x_{23}=|c_{0}|\)), the binary operators \(\{+,-,\times,/\}\), and a complexity cap of ten nodes, over forty evolutionary iterations on a sample of one thousand matrices.

The Pareto front contained four candidate formulas; the one with the best trade-off between error and length was
\begin{equation}
\hat p(A)=0.9720+\frac{0.0331}{1.5541-|c_{0}|},
\end{equation}
an expression that depends solely on the determinant magnitude.  A second-ranked model
uses only the magnitude of the coefficient related to the trace of the adjugate matrix, \(|c_1|\). Classifying a matrix as monotone whenever \(\hat p(A)>0\) reduces, for the leading formula, to the simple scalar condition \(|c_{0}|\lesssim1.59\).   The outcome echoes the decision-tree bands of section~\ref{sec:dt}: under the uniform \((-1,1)\) entry distribution almost the entire monotonicity signal resides in the lowest‐order characteristic coefficient, with \(|c_{1}|\) offering only a modest refinement. We revisit the circle of ideas in section~\ref{sec:reduction}, where we ask how
far one can push a \emph{one-feature} necessary or sufficient condition.

\subsection{The two-feature model}\label{sec:twofeat}

Salience analysis, decision tree surrogates, and symbolic regression all strongly indicated that the absolute values of the lowest-order characteristic polynomial coefficients, \(|c_0|\) and \(|c_1|\), were the dominant features used by our high-accuracy hybrid FFN model. This prompted the question: are these two features alone sufficient for reasonably high classification accuracy?

To verify, we trained a new FFN classifier for the \(n=7\) dataset using \emph{only} \(|c_0|\) and \(|c_1|\) as input. To ensure performance differences primarily reflected feature information content rather than architectural changes, the FFN architecture (hidden layers, activations, regularization) and training procedure were kept identical to the hybrid model (Section~\ref{sec:derivedfeat}), despite the input dimension reducing from 73 to 2.

\begin{figure}[htbp]
  \centering
  \includegraphics[width=0.8\textwidth]{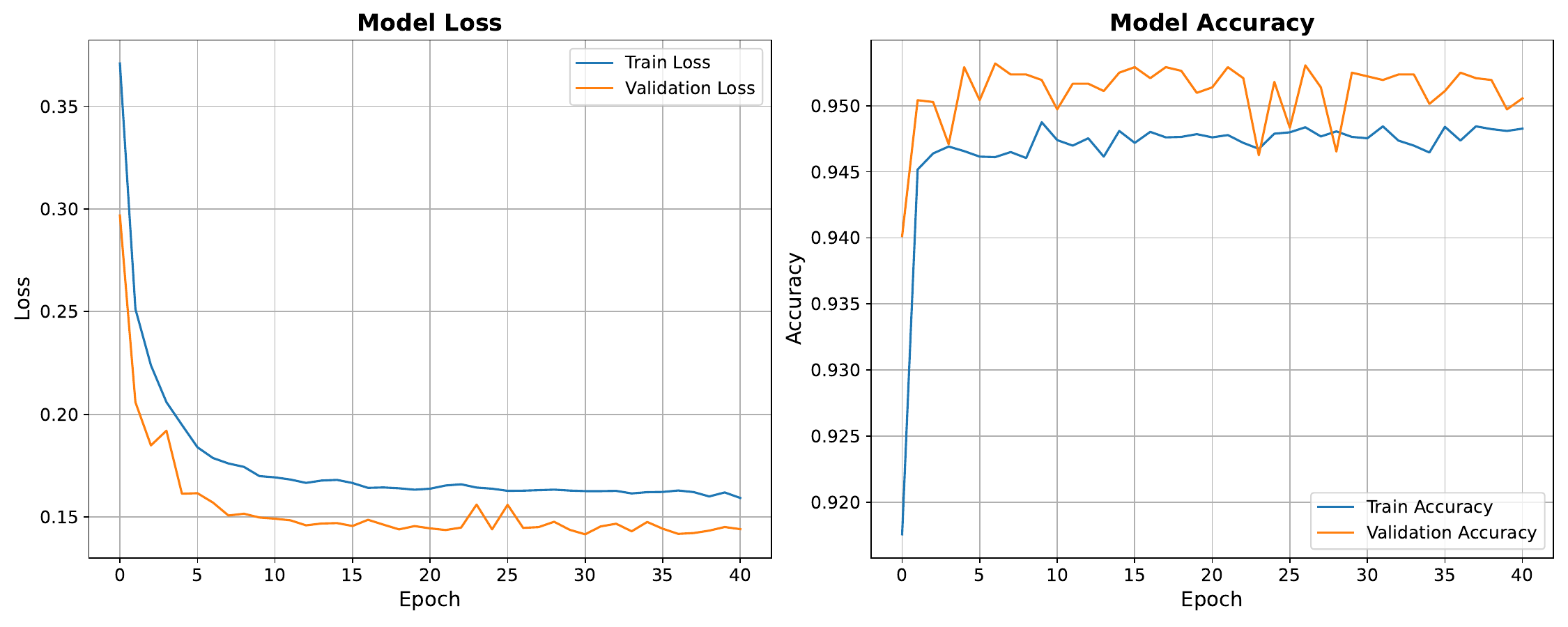}
  \caption{Evolution of training and validation loss and accuracy over epochs for the 7$\times$7 FFN matrix classifier using only the two most salient features: \(|c_0|\) and \(|c_1|\).}
  \label{fig:perf-2feat}
\end{figure}

Figure~\ref{fig:perf-2feat} shows this two-feature model's training evolution. It converged quickly, achieving $\approx 95\%$ validation accuracy. Notably, compared to the 73-feature hybrid model (Fig.~\ref{fig:perf-derived}), this simpler model exhibited significantly less overfitting, with training and validation loss/accuracy curves remaining closer. This suggests that while \(|c_0|\) and \(|c_1|\) alone may not capture enough information for the absolute peak accuracy of the 73-feature model, they form the basis of a more robust model with better generalization on this dataset.

%
%
%

Evaluating this two-feature FFN on the full dataset ($36,000$ matrices) yielded the confusion matrix $\begin{bsmallmatrix} 16708 & 1292\\ 480 & 17520 \end{bsmallmatrix}$.
The corresponding performance metrics are in Table~\ref{tab:dnn-metrics-2feat}.

%
%
%
\begin{table}[htbp]
  \centering
  \caption{Performance of FFN classifier on the 7$\times$7 dataset using only the two most salient features \(|c_0|\) and \(|c_1|\) identified via Integrated Gradients.}
  \label{tab:dnn-metrics-2feat}
  \begin{tabular}{@{}lcccc@{}}
    \toprule
    Model Input & Accuracy & Precision & Recall & Specificity \\
    \midrule
     \(|c_0|\), \(|c_1|\) only (FFN) & 95.1\% & 93.1\% & 97.3\% & 92.8\% \\ 
    \bottomrule
  \end{tabular}
\end{table}

The two-feature model achieved 95.1\% accuracy. While lower than the 99.0\% from the 73-feature hybrid model (Table~\ref{tab:dnn-metrics-derived}), indicating other features contribute predictive information (especially for reducing false positives), this 95.1\% accuracy is still remarkably high and surpasses the baseline CNN (Table~\ref{tab:dnn-metrics-split}).

This strongly validates earlier findings: \(|c_0|\) and \(|c_1|\) magnitudes capture most of the signal needed to distinguish monotone from non-monotone matrices in this dataset, yielding high accuracy independently. The $\approx 4\%$ accuracy drop versus the hybrid model quantifies the remaining features' contribution to refining the boundary, particularly for ambiguous cases signaled by \(|c_0|, |c_1|\).

\section{Further search for separation} \label{sec:separation_search}

Previous sections highlighted the dominance of \(|c_0|\) and \(|c_1|\) in classifying monotonicity via FFNs, though scatter plots (Fig.~\ref{fig:scatter_c0_c1}) showed an imperfect, non-linear boundary. This motivated exploring explicit mathematical separators using Support Vector Machines (SVMs)~\cite{cortes1995support}.

\begin{figure}[htbp]
  \centering
  \includegraphics[width=0.75\textwidth]{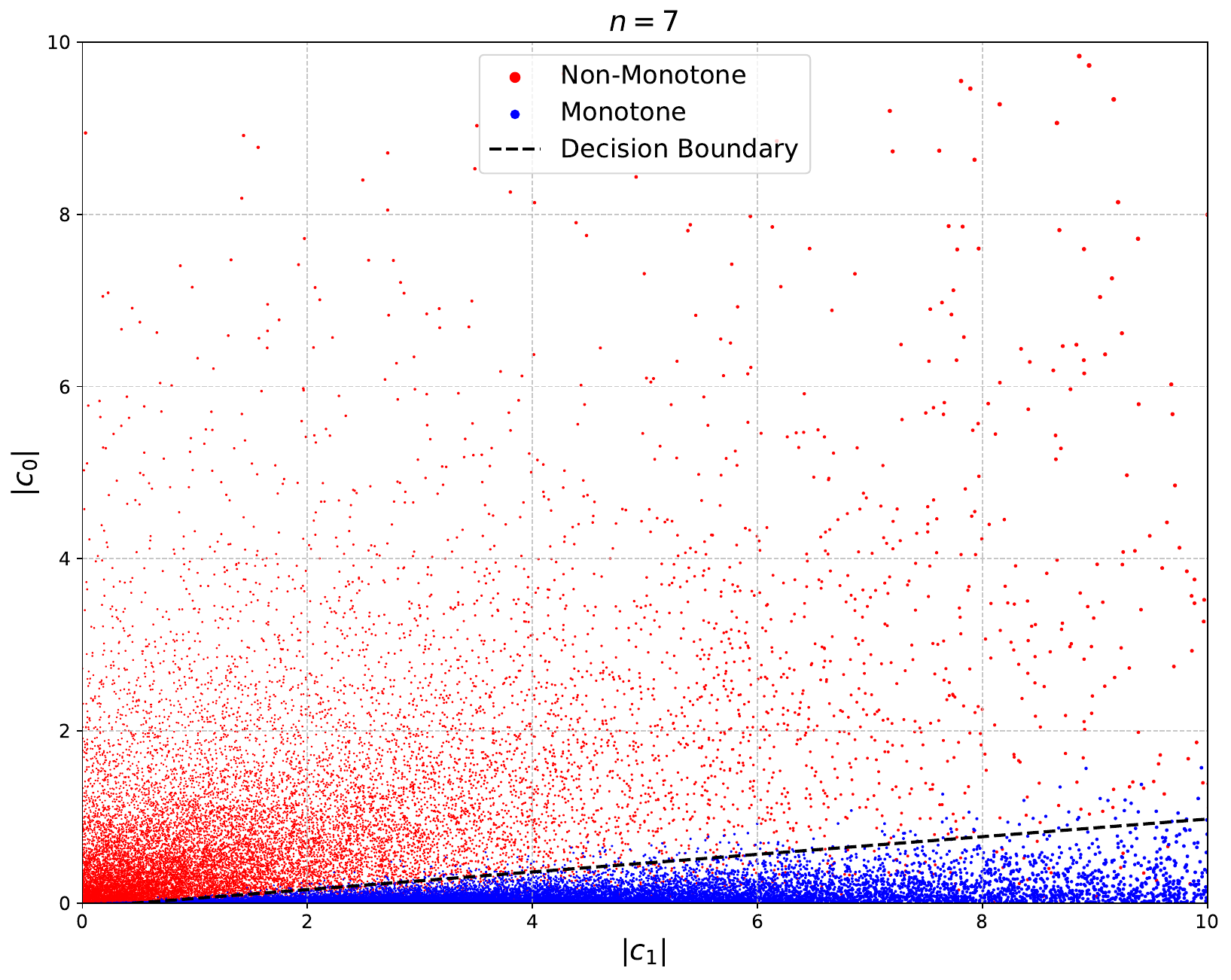}
  \caption{Scatter plot of \(|c_0|\) vs \(|c_1|\) for \(n=7\) matrices, showing monotone (blue) and non-monotone (red) points. The dashed line represents the linear separation boundary found by the SVM with a linear kernel (Equation~\ref{eq:svm_linear_2feat}).}
  \label{fig:svm_scatter_7}
\end{figure}

With \(|c_0|\) and \(|c_1|\) as features, SVMs with linear, polynomial (degrees 3/5, $C=0.5$), and RBF kernels yielded $\approx 94.6\%$ accuracy, comparable to the 95.1\% of the two-feature FFN (Table~\ref{tab:dnn-metrics-2feat}); the sigmoid kernel performed poorly (59.1\%). The linear SVM separating hyperplane (Eq.~\ref{eq:svm_linear_2feat}; plotted in Fig.~\ref{fig:svm_scatter_7}) is
\begin{equation} \label{eq:svm_linear_2feat}
 -12.38\,|c_0| + 1.27\,|c_1| - 0.59 = 0.
\end{equation}
While roughly separating classes, it misclassifies near-origin points where classes overlap. Symbolic regression (PySR) suggested non-linear boundaries with logarithmic terms in \(|c_0|, |c_1|\) but offered no significant accuracy improvement over this linear SVM, suggesting fundamental class overlap limits gains with just these two features.

%
%

We also explored SVM performance using all seven absolute characteristic polynomial coefficients (\(|c_0|, \dots, |c_6|\), from DF7). This linear SVM achieved $94.6\%$ accuracy (Confusion Matrix: $\begin{bsmallmatrix} 16559 & 1441 \\ 510 & 17490 \end{bsmallmatrix}$), negligibly different from using only \(|c_0|\) and \(|c_1|\). The resulting 7-dimensional hyperplane equation is:
\begin{equation} \label{eq:svm_linear_7feat}
\begin{split}
    -0.0068\,|c_6| + 0.0067\,|c_5| - 0.0135\,|c_4| + 0.0045\,|c_3| \\
    - 0.0104\,|c_2| + 1.2782\,|c_1| - 12.4071\,|c_0| - 0.5675 = 0.
\end{split}
\end{equation}
Comparing coefficients in \eqref{eq:svm_linear_7feat} with \eqref{eq:svm_linear_2feat}, the SVM still overwhelmingly weights \(|c_0|\) (coeff. $\approx -12.4$) and \(|c_1|\) (coeff. $\approx 1.3$), while coefficients for \(|c_2|\) through \(|c_6|\) are orders of magnitude smaller. This confirms that for linear separation, adding higher-order coefficient magnitudes offers minimal benefit beyond \(|c_0|,|c_1|\). While non-linear models like the hybrid FFN utilize the full feature set for higher accuracy, this search for a simple, interpretable linear boundary reinforces \(|c_0|\) and \(|c_1|\) as the decisive parameters.

\section{Reduction of one explaining feature}\label{sec:reduction}

Our previous results point to $|c_0|$ and $|c_1|$ as the key indicators of whether a matrix is monotone.
While the two-feature model (section~\ref{sec:twofeat}) achieved remarkable 95.1\% accuracy and good generalization, the distinct visual patterns in the \((|c_1|, |c_0|)\) plane (Figure~\ref{fig:scatter_c0_c1}) is clear: monotonicity appears associated with small $|c_0|$ {\em and} larger $|c_1|$, suggesting that smaller values of $| \frac{c_0}{c_1}  |$  would indicate monotone matrices. This invites further investigation into simpler, potentially one-dimensional criteria. Could a single quantity derived from these two coefficients capture a significant portion of the decision boundary?

A natural candidate is the ratio of the two dominant coefficient magnitudes, \(|c_0/c_1|\) (corresponding to feature DF8 in Table~\ref{tab:derived-features-full}, which was set aside earlier). This ratio essentially measures how small the determinant magnitude is relative to the \(|c_1|\) coefficient magnitude.

Beyond the empirical motivation, this specific ratio possesses a neat mathematical connection to the matrix inverse, which is central to the definition of monotonicity. Recalling that \(c_0 = (-1)^n \det(A)\) and \(c_1 = (-1)^{n-1} \mathrm{tr}(\mathrm{adj}(A))\), and using the identity \(\mathrm{adj}(A) = \det(A) A^{-1}\) for an invertible matrix \(A\), we have \(c_1 = (-1)^{n-1} \det(A) \mathrm{tr}(A^{-1})\). Therefore, provided \(c_1 \neq 0\) (or equivalently, \(\mathrm{tr}(A^{-1}) \neq 0\)), the ratio is
\[
\frac{c_0}{c_1} = \frac{(-1)^n \det(A)}{(-1)^{n-1} \det(A) \mathrm{tr}(A^{-1})} = -\frac{1}{\mathrm{tr}(A^{-1})}\,.
\]
Taking the magnitude gives
\begin{equation} \label{eq:ratio_trace_inv}
\left| \frac{c_0}{c_1} \right| = \frac{1}{|\mathrm{tr}(A^{-1})|} \,.
\end{equation}
For a general invertible matrix, the trace of the inverse can be positive, negative, or zero (in which case the ratio is undefined). However, for a \emph{monotone} matrix \(A\), we know that \(A^{-1}\ge 0\) and is not the zero matrix, which strictly requires \(\mathrm{tr}(A^{-1}) = \sum_i (A^{-1})_{ii} > 0\). In this specific case, the relation simplifies to:
\begin{equation} \label{eq:ratio_trace_inv_mono}
\left| \frac{c_0}{c_1} \right| = \frac{1}{\mathrm{tr}(A^{-1})} \quad (\text{for } A \text{ monotone}).
\end{equation}
This identity offers a clear theoretical basis for the observed importance of the ratio \(|c_0/c_1|\). It directly links small values of this empirically significant ratio to large positive values of the trace of the inverse matrix, a property intrinsically related to the non-negativity required by monotonicity (\(A^{-1}\ge0\)).


\subsection{\texorpdfstring{Classification using the ratio $|c_0/c_1|$}{Classification using the ratio |c0/c1|}}\label{sec:ratio_classification}

To examine the predictive power of this single ratio, we trained an FFN classifier using only \(|c_0/c_1|\) as input. As in the previous section, to isolate the effect of the feature itself, we maintained the identical FFN architecture (hidden layers, activations, regularization) used for the hybrid and two-feature models, adapting only the input layer dimension to one. The training process also remained unchanged.

\begin{figure}[htbp]
  \centering
  \includegraphics[width=0.8\textwidth]{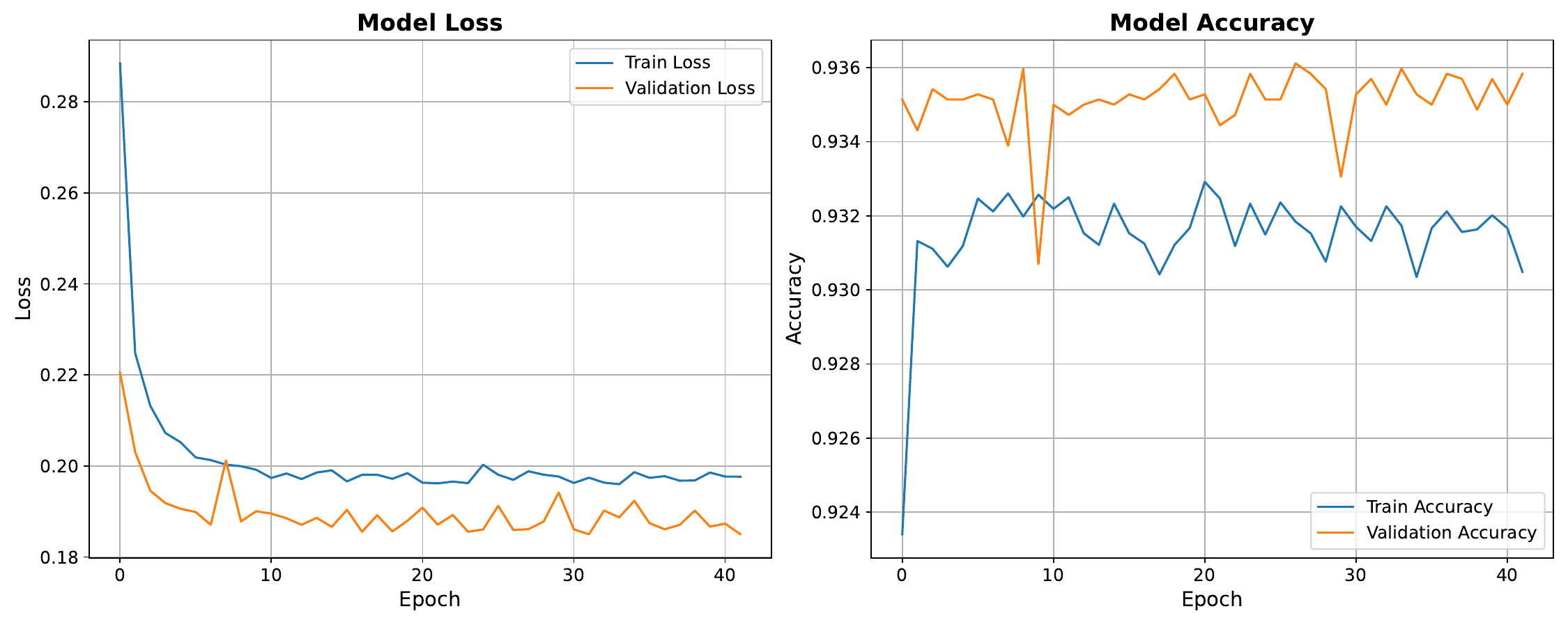}
  \caption{Evolution of training and validation loss and accuracy over epochs for the 7$\times$7 FFN matrix classifier using only the single feature \(|c_0 / c_1|\).}
  \label{fig:perf-ratio}
\end{figure}

Figure~\ref{fig:perf-ratio} displays the training evolution for this single-feature (ratio) model. The model achieves a validation accuracy fluctuating mostly between 93\% and 94\%. A remarkable aspect of these curves is the relationship between training and validation performance. Unlike the hybrid model which showed hints of overfitting (Figure~\ref{fig:perf-derived}), here the validation accuracy almost always exceeds the training accuracy, and the validation loss remains  below the training loss. This pattern, where the model performs as well or better on unseen validation data as it does on the training data,  indicates a strong  generalization and was observed for the $n=5$ and $n=10$ cases as well. It suggests that the single feature \(|c_0/c_1|\), while perhaps insufficient to capture the full complexity needed for $>99\%$ accuracy, provides a very robust signal that prevents the complex FFN architecture from merely memorizing the training set. Given the balanced dataset and lack of data leakage, this superior generalization is a significant finding.

%
%
%

Evaluating this single-feature FFN model on the full dataset (36,000 matrices) resulted in the following confusion matrix:
\[
\begin{bmatrix}
\mathrm{TN} & \mathrm{FP}\\
\mathrm{FN} & \mathrm{TP}
\end{bmatrix}_{\text{FFN-Ratio}}
=
\begin{bmatrix}
16041 & 1959\\ 
396   & 17604
\end{bmatrix}.
\]
Table~\ref{tab:dnn-metrics-ratio} summarizes the corresponding performance metrics.

\begin{table}[htbp]
  \centering
  \caption{Performance of FFN classifier on the 7$\times$7 dataset using only the single derived feature ratio \(|c_0 / c_1|\).}
  \label{tab:dnn-metrics-ratio}
  \begin{tabular}{@{}lcccc@{}}
    \toprule
    Model Input & Accuracy & Precision & Recall & Specificity \\
    \midrule
     \(|c_0 / c_1|\) only (FFN) & 93.5\% & 90.0\% & 97.8\% & 89.1\% \\
    \bottomrule
  \end{tabular}
\end{table}

The overall accuracy using only the ratio \(|c_0 / c_1|\) is 93.5\%. While this is lower than the 95.1\% achieved by the two-feature model (Table~\ref{tab:dnn-metrics-2feat}), it remains a relatively high value, confirming that this single ratio indeed carries significant predictive information about monotonicity within this dataset. The decrease in accuracy primarily manifests as lower precision and specificity, indicating more false positives compared to the two-feature model. 

Aiming at the simplest possible rule based on this single feature, we trained a CART-style decision tree (depth 1) directly on the \(n=7\) dataset using only \(|c_0 / c_1|\) as input. The resulting optimal split, maximizing overall classification accuracy, yielded what we term the {\em Decision Tree Threshold}, denoted \(T_{\text{DT},7}\). For our data, this was found to be:
\[
T_{\text{DT},7} = 0.08,
\]
meaning the simple rule: Predict Monotone if \(|c_0 / c_1| \le T_{\text{DT},7}\), and Non-monotone if \(|c_0 / c_1| > T_{\text{DT},7}\). Interestingly, as we will discuss in Section~\ref{sec:necessary} when analyzing the tail behavior of the monotone class, this value \(T_{\text{DT},7}\) appears to coincide with a region where modeling the extremes becomes less stable.

However, a threshold optimized for overall accuracy like \(T_{\text{DT},7}\) inevitably misclassifies some instances; in particular, it may classify some matrices as monotone even if their ratio exceeds this value. The purpose of obtaining a potential \emph{necessary condition} for monotonicity based on this dataset, made us search the maximum observed value of the ratio \(|c_0 / c_1|\) among only the monotone matrices in our dataset. We term this value the {\em Sample Maximum Ratio}, denoted simply as \(T_n\) for each matrix size \(n\).

\begin{figure}[htbp]
  \centering
  \subfloat[\(T_5 = 0.5968\)]{\label{fig:dist_n5}\includegraphics[width=0.325\textwidth]{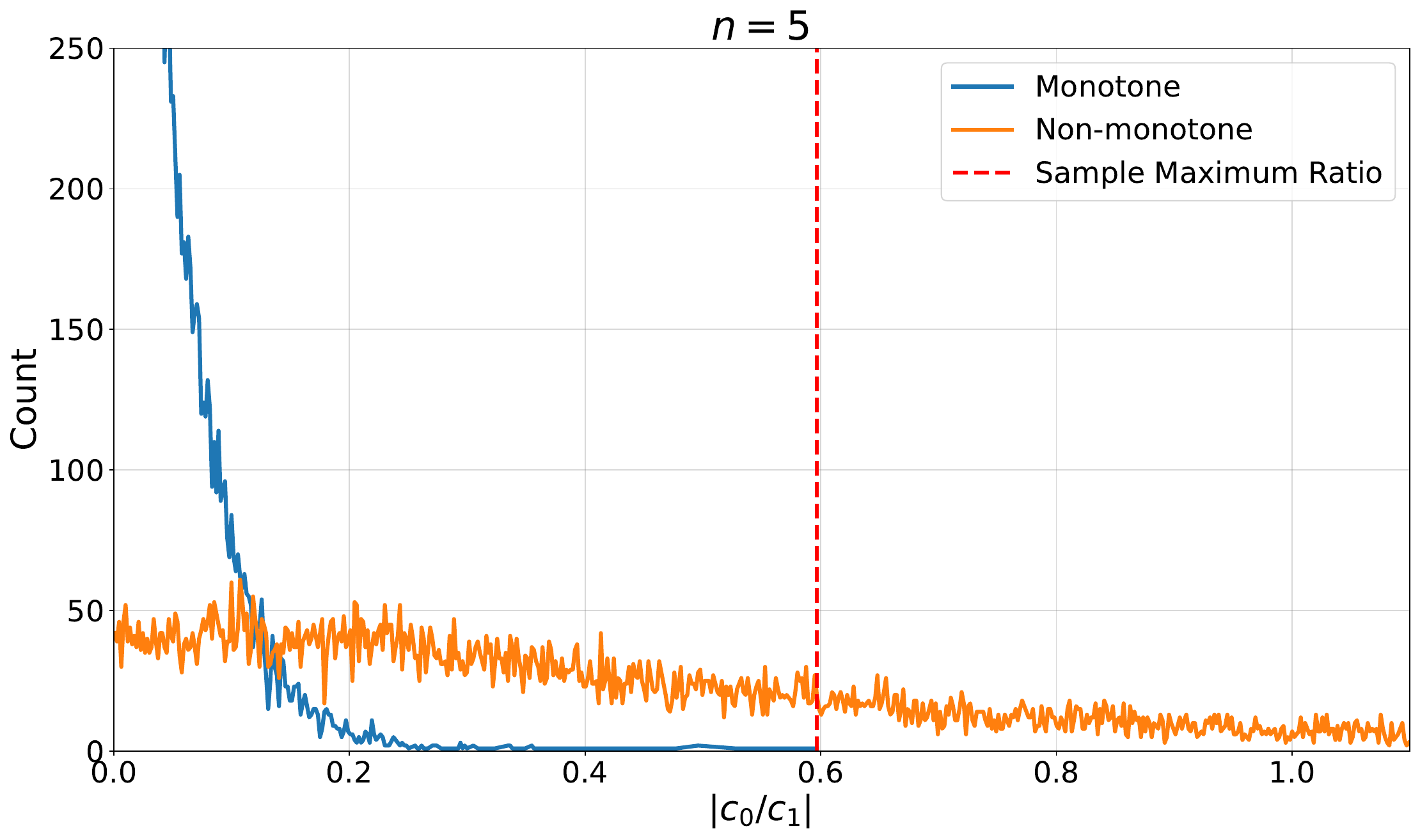}}
  \hfill 
  \subfloat[\(T_7 = 0.1755\)]{\label{fig:dist_n7}\includegraphics[width=0.325\textwidth]{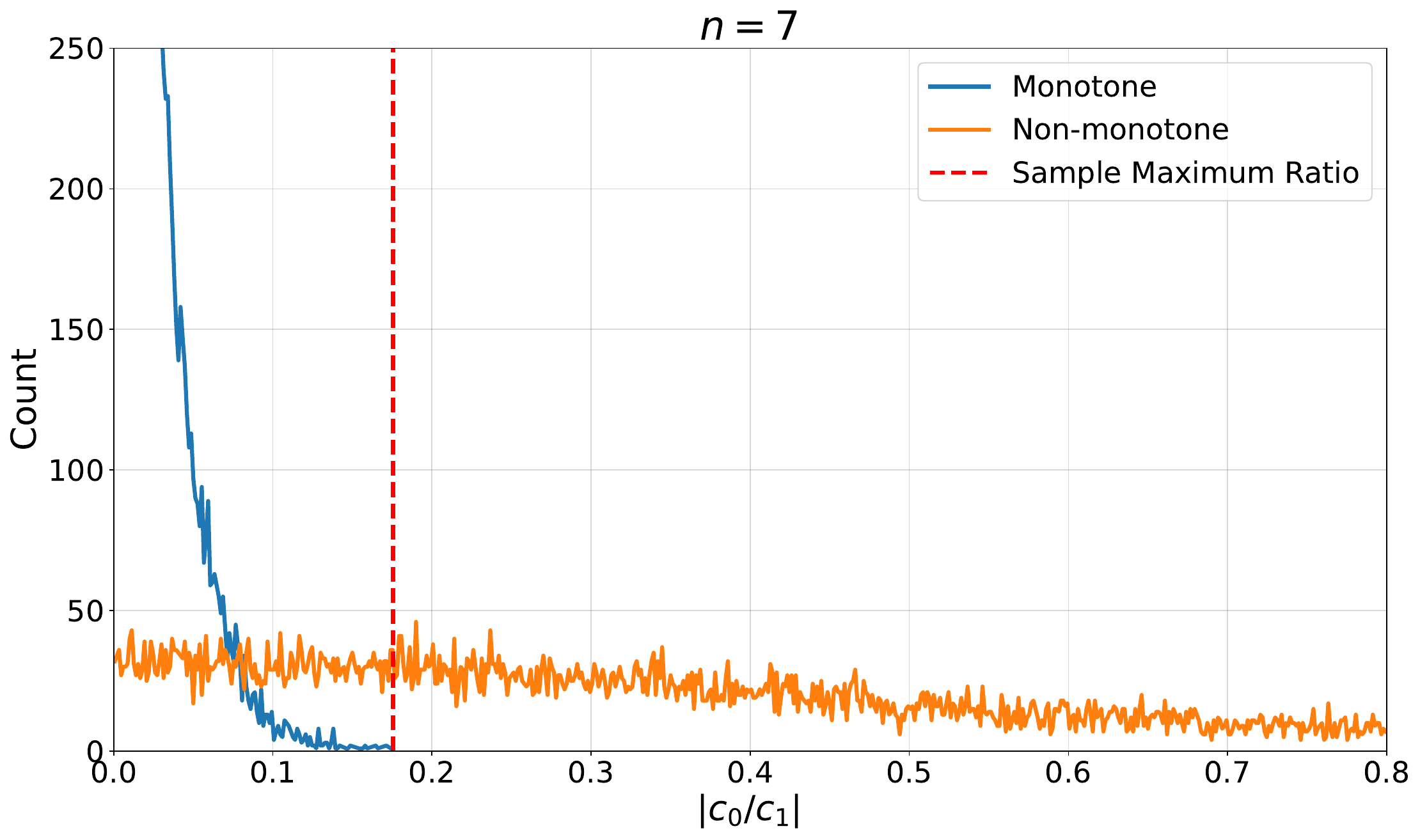}}
  \hfill 
  \subfloat[\(T_{10} = 0.104\)]{\label{fig:dist_n10}\includegraphics[width=0.325\textwidth]{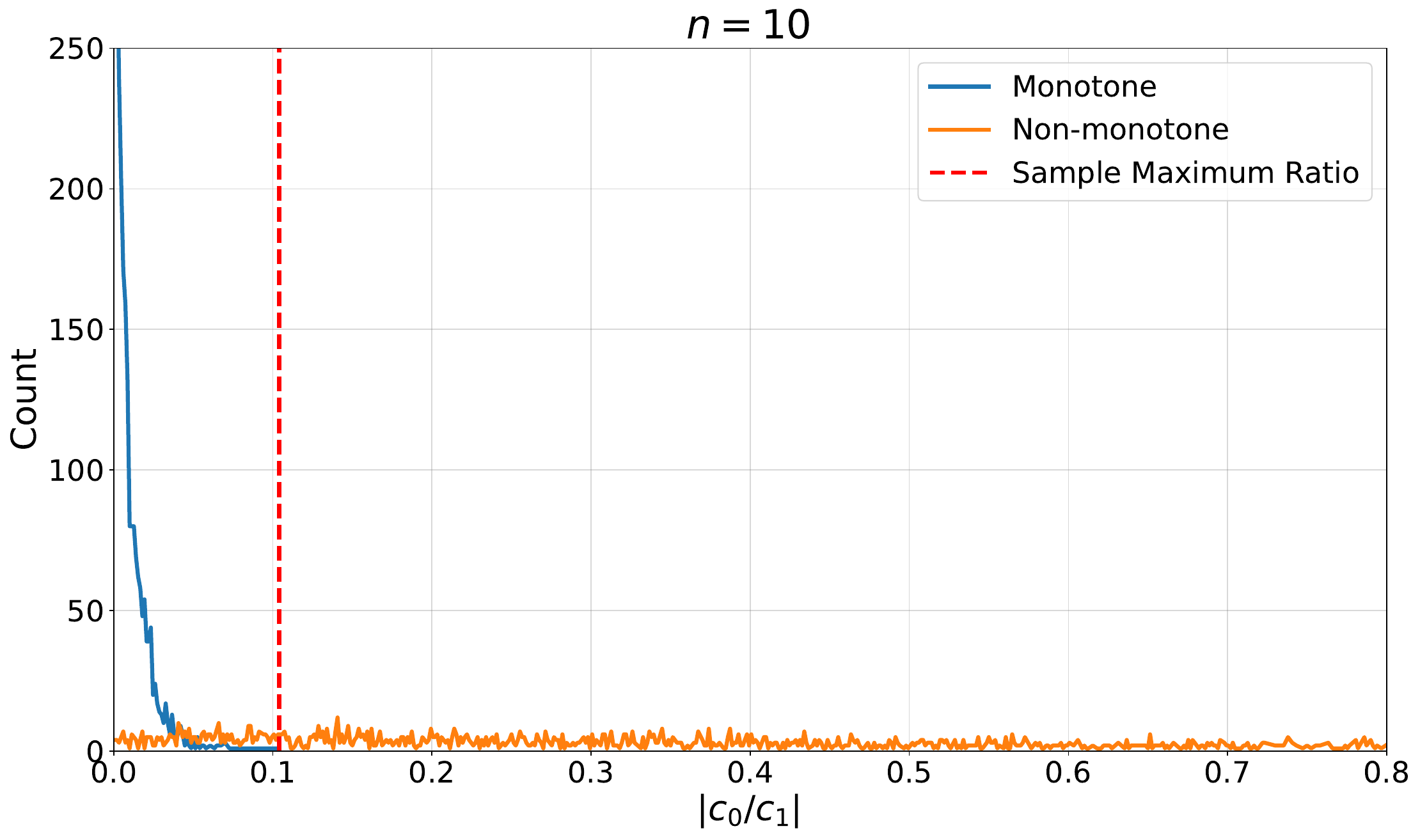}}
\caption{Line plots of frequency counts showing the distribution of the feature ratio \(|c_0 / c_1|\) for monotone (blue) and non-monotone (orange) matrices for \(n=5, 7,\) and \(10\). The vertical dashed red line indicates the {\em Sample Maximum Ratio} (\(T_n\)) for monotone matrices in each dataset, which forms the basis for a potential necessary condition threshold.}
\label{fig:ratio_distributions}
\end{figure}

To visualize the distribution of the ratio for both classes and identify these \(T_n\) values empirically, Figure~\ref{fig:ratio_distributions} shows line plots of frequency counts of \(|c_0 / c_1|\) for monotone (blue) and non-monotone (orange) matrices across the different dataset sizes (\(n=5, 7, 10\)). A vertical dashed red line indicates the maximum observed value of \(|c_0 / c_1|\) for the monotone matrices in each respective dataset.

%
%

As clearly seen in Figure~\ref{fig:ratio_distributions}, the monotone matrices consistently exhibit much smaller values of the ratio \(|c_0 / c_1|\) compared to the non-monotone matrices.  Furthermore, for each matrix size \(n\), there exists a maximum observed value of this ratio among all monotone matrices in the dataset.  These {\em Sample Maximum Ratios}, \(T_5 = 0.5968\), \(T_7 = 0.1755\), and \(T_{10} = 0.104\), define statistically derived necessary conditions based on our data: any matrix \(A\) in the respective dataset found to be monotone satisfied \(|c_0 / c_1| \le T_n\). This observation forms the basis for the condition presented in the  next subsection.
%
\subsection{Necessary Condition for Monotonicity} \label{sec:necessary}
The results from the previous subsection, particularly the lines of frequency counts in Figure~\ref{fig:ratio_distributions}, made clear, for all matrix sizes $n$ studied, an empirical maximum value, \(T_n\), such that all monotone matrices in our  dataset satisfied \(r \le T_n\). While this empirical observation is strong, to establish a more rigorous basis for a necessary condition, we use methods from Extreme Value Theory (EVT) \cite{coles2001introduction}.

EVT provides a framework for modeling the tail behavior of probability distributions. Specifically, the Peaks-Over-Threshold (POT) approach assumes that for a sufficiently high threshold \(u\), the distribution of the excesses (\(X-u\)) for values \(X\) exceeding \(u\) can be well approximated by the Generalized Pareto Distribution (GPD). The GPD is characterized by a shape parameter \(\xi\) and a scale parameter \(\sigma\). The sign of the shape parameter \(\xi\) tells us what kind of tail the distribution has: \(\xi < 0\) implies the distribution has a finite upper endpoint, \(\xi = 0\) corresponds to an exponential-like tail (e.g., Gaussian), and \(\xi > 0\) indicates a heavy tail with no upper bound (e.g., Pareto).

\begin{figure}[htbp]
  \centering
  \subfloat[Mean Residual Life Plot]{\label{fig:mrl_plot_n7}\includegraphics[width=0.55\textwidth]{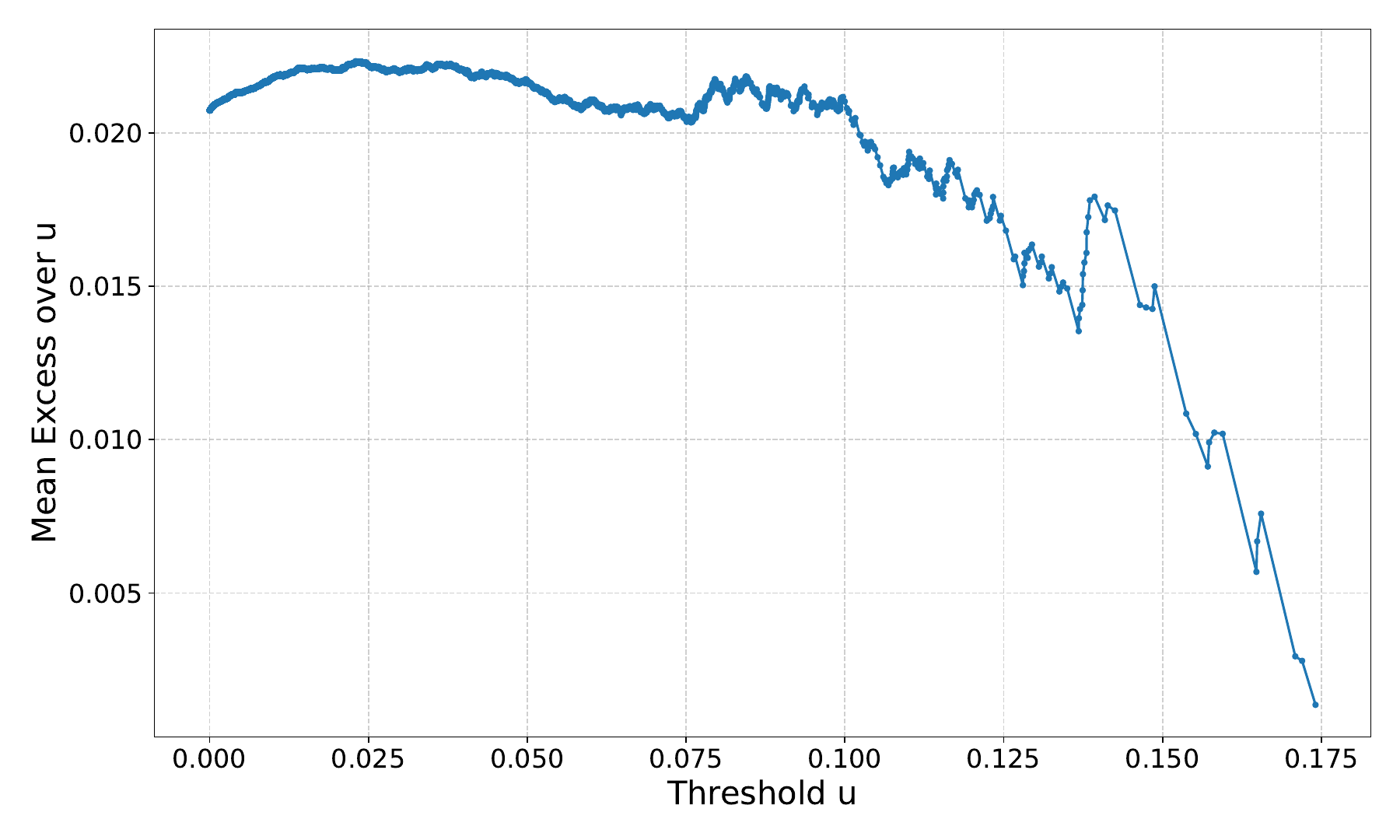}}
  \hfill
  \subfloat[GPD Parameter Stability (\(\xi\), top; \(\sigma\), bottom)]{\label{fig:param_stability_n7}\includegraphics[width=0.44\textwidth]{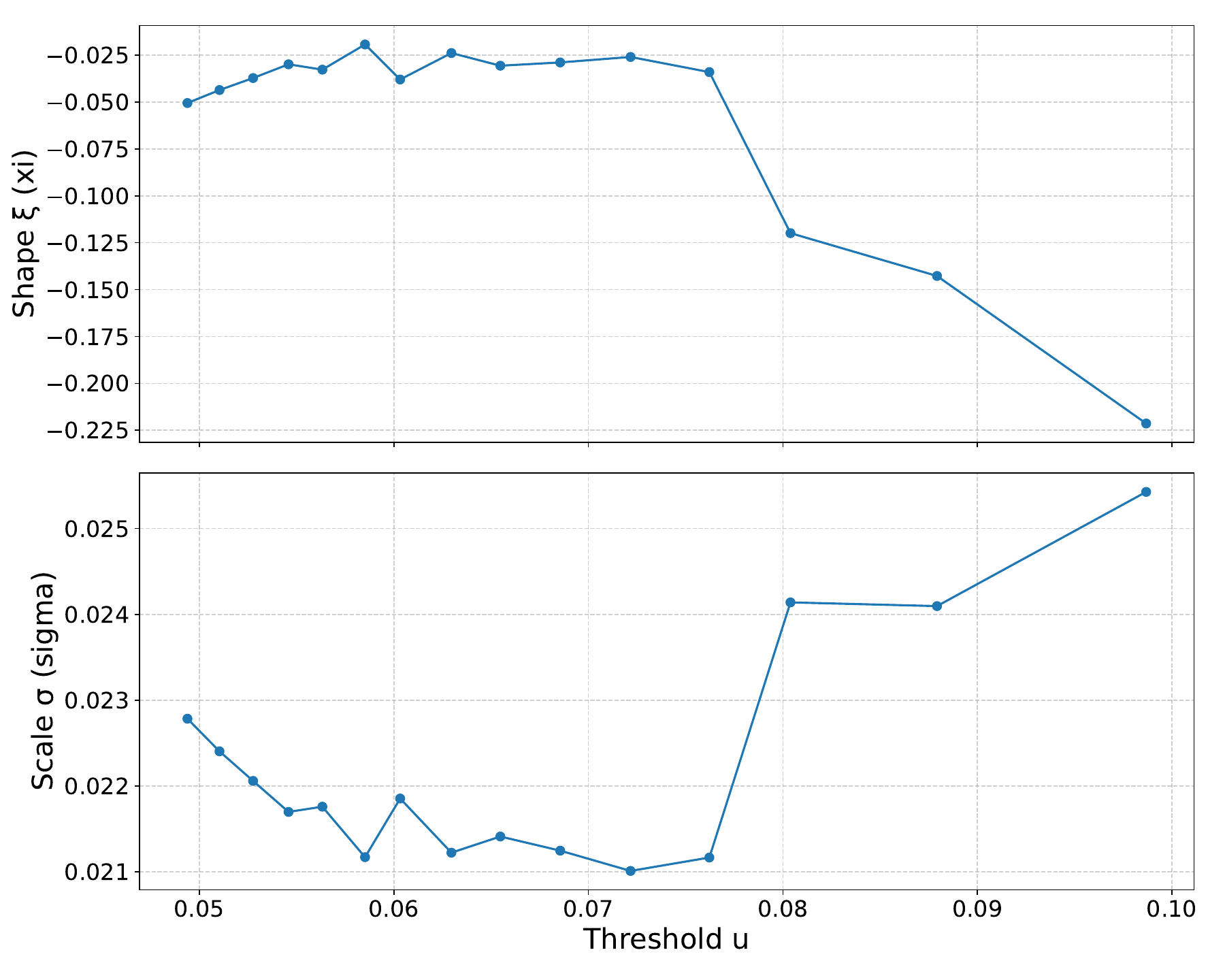}}
  \caption{EVT diagnostic plots for the ratio \(|c_0 / c_1|\) using the \(n=7\) monotone matrix dataset. (a) The downward trend in the MRL plot for \(u \gtrsim 0.1\) suggests \(\xi < 0\). (b) The stability plot shows relatively stable negative \(\xi\) estimates for \(u\) between approximately 0.05 and 0.075, before becoming unstable.}
  \label{fig:evt_diagnostics_n7}
\end{figure}

We applied the POT method to the distribution of the ratio \(r = |c_0/c_1|\) for the monotone matrices in our dataset, focusing on the \(n=7\) case. Standard EVT diagnostic plots are used to guide the choice of an appropriate threshold \(u\). The Mean Residual Life (MRL) plot (Figure~\ref{fig:mrl_plot_n7}) should exhibit approximate linearity above \(u\), with a downward slope indicating \(\xi < 0\). The parameter stability plots (Figure~\ref{fig:param_stability_n7}) show the estimated \(\xi\) and \(\sigma\) as \(u\) varies; we seek a range where these estimates are reasonably stable.

%
%
%

Based on the relative stability in Figure~\ref{fig:param_stability_n7}(b) for \(u \in [0.06,0.075]\) and the mean residual life trend, we fixed \(u = 0.075\).  Fitting a generalized Pareto distribution to the \(N_u = 562\) excesses above this threshold yields
\[
  \hat{\xi} = -0.028 \pm 0.006,\qquad
  \hat{\sigma} = 0.021 \pm 0.003
  \quad(\text{MLE} \pm \text{s.e.}).
\]
The significantly negative \(\hat{\xi}\) provides strong statistical evidence for a finite upper endpoint, which we estimate as
\[
  E_7 \;=\; u \;-\;\frac{\hat{\sigma}}{\hat{\xi}}
  \;\approx\; 0.828.
\]
We can use the fitted GPD model to estimate the probability of exceeding the maximum ratio observed in the data, \(T_7 \approx 0.1755\). The model yields 
\[
P(r > 0.1754 \, | \, \text{monotone}, n=7) \approx 1.8 \times 10^{-4}.
\]
To assess the robustness of this estimate, a 95\% bootstrap confidence interval was computed, yielding \([6.8 \times 10^{-5}, 3.1 \times 10^{-5}]\), and a threshold stability scan over \(u \in [0.049, 0.083]\) yielded a mean probability of \((1.61 \pm 0.26) \times 10^{-4}\). Both analyses confirm that the probability of exceeding the observed maximum is robustly estimated as negligible.

Intriguingly, the threshold region \(u \approx 0.08\) where the GPD parameter estimates begin to show instability in Figure~\ref{fig:param_stability_n7}(b), aligns remarkably well with the decision tree threshold \(T_{\text{DT},7} \approx 0.08\) identified in Section~\ref{sec:reduction}. This suggests a potential connection between the limits of class separability based on this single ratio and the onset of complex tail behavior within the monotone class itself.

Similar EVT analyses were performed for \(n=5\) and \(n=10\). Interestingly, these yielded positive shape parameters (\(\xi_5 \approx 0.077\) for \(u=T_{\text{DT},5} = 0.13\); \(\xi_{10} \approx 0.037\) for \(u=T_{\text{DT},10} = 0.04\)), suggesting heavy tails {\em without} a finite endpoint according to the GPD model in those cases. The decision tree thresholds, \(T_{\text{DT},5} = 0.13\) (\(n=5\)) and \(T_{\text{DT},10} = 0.04\) (\(n=10\)), found using the same decision tree methodology, again corresponded roughly to regions where GPD parameter stability deteriorated in their respective analyses (plots not shown). While the lack of a negative \(\xi\) for \(n=5\) and \(n=10\) means the finite endpoint argument does not  universally hold across these fits, the clear sample maximum ratio \(T_n\) observed for all sizes (Figure~\ref{fig:ratio_distributions}) and the compelling statistical evidence supporting this limit for \(n=7\) (\(\xi < 0\), negligible exceedance probability) leads us to formulate the following data-driven necessary condition specifically for the case mainly studied.
\bigskip

\noindent\textbf{Proposed Necessary Condition (Statistical Bound for \(n=7\)).}
\textit{Let \(A \in \mathbb{R}^{7\times 7}\) have independent entries sampled uniformly from \((-1,1)\). If \(A\) is monotone (\(A^{-1} \ge 0\)), then it is proposed that the following condition holds with high statistical probability\footnote{This condition is based on \(18{,}000\) observed monotone samples, none of which violated it (\(T_7 \approx 0.1755\)). 
Extreme-value analysis estimates the exceedance probability \(P\!\bigl(|c_0/c_1| > 0.1755 \,\big|\, \text{monotone},\, n=7\bigr) \approx (1.6 \pm 0.3)\times10^{-4}\) and shows that this value is stable across thresholds \(u\in[0.049,0.083]\). 
For a representative threshold \(u=0.075\), a 95\% bootstrap confidence interval for the same probability is \([6.8\times10^{-5},\, 3.1\times10^{-4}]\). 
While these results strongly support the condition, it remains an empirically derived statistical bound, not a formally proven theorem; its applicability to other matrix sizes or entry distributions requires separate investigation.}:
\begin{equation} \label{eq:necessary_condition_n7}
    \frac{1}{\mathrm{tr}(A^{-1})} = \left| \frac{c_0}{c_1} \right| \le 0.1755 \,,
\end{equation}
where \(c_0 = \det(-A)\) and \(c_1 = (-1)^{6} \mathrm{tr}(\mathrm{adj}(A))\).}
\bigskip
\begin{remark}[Conjectured Finite Endpoint (\(n=7\)) and Empirical Context] \label{rem:endpoint_conjecture_context}
This proposed condition offers a simple, statistically robust necessary test for monotonicity for \(n=7\) matrices drawn from the specified uniform distribution, directly in terms of the trace of the non-negative inverse matrix \(A^{-1}\) (or equivalently, via the ratio \(|c_0/c_1|\)). It implies that for such monotone matrices, \(\mathrm{tr}(A^{-1})\) is likely bounded below by \(1/0.1755 \approx 5.7\).
While similar Sample Maximum Ratios \(T_n\) were observed for other dataset sizes (\(T_5 \approx 0.5968\), \(T_{10} \approx 0.104\); see Figure~\ref{fig:ratio_distributions}), the Extreme Value Theory analysis provided the strongest statistical support for a bounded tail specifically for the \(n=7\) case, where the estimated shape parameter \(\xi\) was negative. This leads us to conjecture that for \(7 \times 7\) monotone matrices with entries uniformly distributed in \((-1,1)\), the distribution of the ratio \(|c_0/c_1|\) indeed possesses a finite upper endpoint. Determining if this endpoint corresponds precisely to the empirically observed \(T_7\) or a slightly larger value (like the EVT-estimated \(E_7 \approx 0.828\)), and proving its existence and value, remains a subject for further research.
\end{remark}

\begin{remark}[Utility Beyond Computational Speed] \label{rem:utility}
It is important to note that the principal value of the proposed necessary condition \eqref{eq:necessary_condition_n7}, based on \(|c_0/c_1|\) (or equivalently \(1/\mathrm{tr}(A^{-1})\)), is not in offering a computationally cheaper alternative for verifying the monotonicity of an individual matrix. The calculation of the characteristic polynomial coefficients \(c_0\) and \(c_1\) generally carries a computational complexity (typically \(O(n^3)\)) comparable to that of direct matrix inversion followed by an element-wise check of \(A^{-1}\ge0\). Rather, its relevance lies in the insight it offers. The consistent emergence of \(|c_0|\) and \(|c_1|\) as dominant features in our AI models (Section~\ref{sec:salience_and_features}) shows that the deep networks primarily uses global algebraic properties encoded in these low-order coefficients to achieve high classification accuracy. This finding points on a simpler, more fundamental understanding of the characteristics that distinguish monotone from non-monotone matrices within the studied distribution, offering a concise explanation for the network learned behavior. The algebraic nature of the condition, compared to the direct definition of monotonicity, may also facilitate further theoretical investigation.
\end{remark}

\begin{remark}[Spectral Invariance and Non-Monotonicity] \label{rem:similarity}
The proposed necessary condition \eqref{eq:necessary_condition_n7} is expressed in terms of the characteristic polynomial coefficients \(c_0\) and \(c_1\). Since these coefficients (and thus their ratio \(|c_0/c_1|\)) are invariant under similarity transformations, the condition offers a spectral perspective. Specifically, if the condition is violated for a given matrix \(A\) (i.e., if \(|c_0(A)/c_1(A)| > 0.1755\)), then, according to our statistical findings, \(A\) is highly unlikely to be monotone. Consequently, any matrix \(B\) similar to \(A\) (\(B = P^{-1}AP\)) will have the same ratio \(|c_0(B)/c_1(B)|\) and would thus also be deemed non-monotone by this criterion. While monotonicity itself is not preserved under general similarity transformations, this observation implies that our data-driven condition, being spectral in nature, identifies a property related to non-monotonicity that is shared by an entire similarity class of matrices.
\end{remark}

\section{Classification and Explanation within a Filtered Subdomain}\label{sec:filtered_subdomain}

The analyses in previous sections, particularly the decision tree (Section~\ref{sec:dt}) and SVM results (Section~\ref{sec:separation_search}), indicated that while the ratio \(|c_0/c_1|\) is a powerful discriminator, the decision boundary is imperfect, especially for matrices where this ratio is small. The Decision Tree Threshold \(T_{\text{DT},7} \approx 0.08\) provided a reasonable first-pass separation. However, within the subdomain \(\Omega_0 = \{A : |c_0/c_1| < T_{\text{DT},7} \}\), a mixture of monotone and non-monotone matrices still exists. This motivates a focused analysis: can we achieve better classification and, more importantly, a clearer explanation for matrices that fall into this ambiguous region, where the principal ratio-based rule might be insufficient?

To address this, we employed a strategy of {\em subdomain classification}. We first filtered the original \(n=7\) dataset to retain only matrices \(A \in \Omega_0\), i.e., those satisfying \(|c_0/c_1| < 0.08\). This {\em zooms in} on the region where the classes are most interlaced according to the global simple rule. Due to the filtering, the class distribution within \(\Omega_0\) became imbalanced. To counteract this, the dataset was subsequently balanced by randomly undersampling the majority class, ensuring an equal number of monotone and non-monotone samples for training the subsequent model. 

\begin{figure}[htbp]
  \centering
  \includegraphics[width=0.8\textwidth]{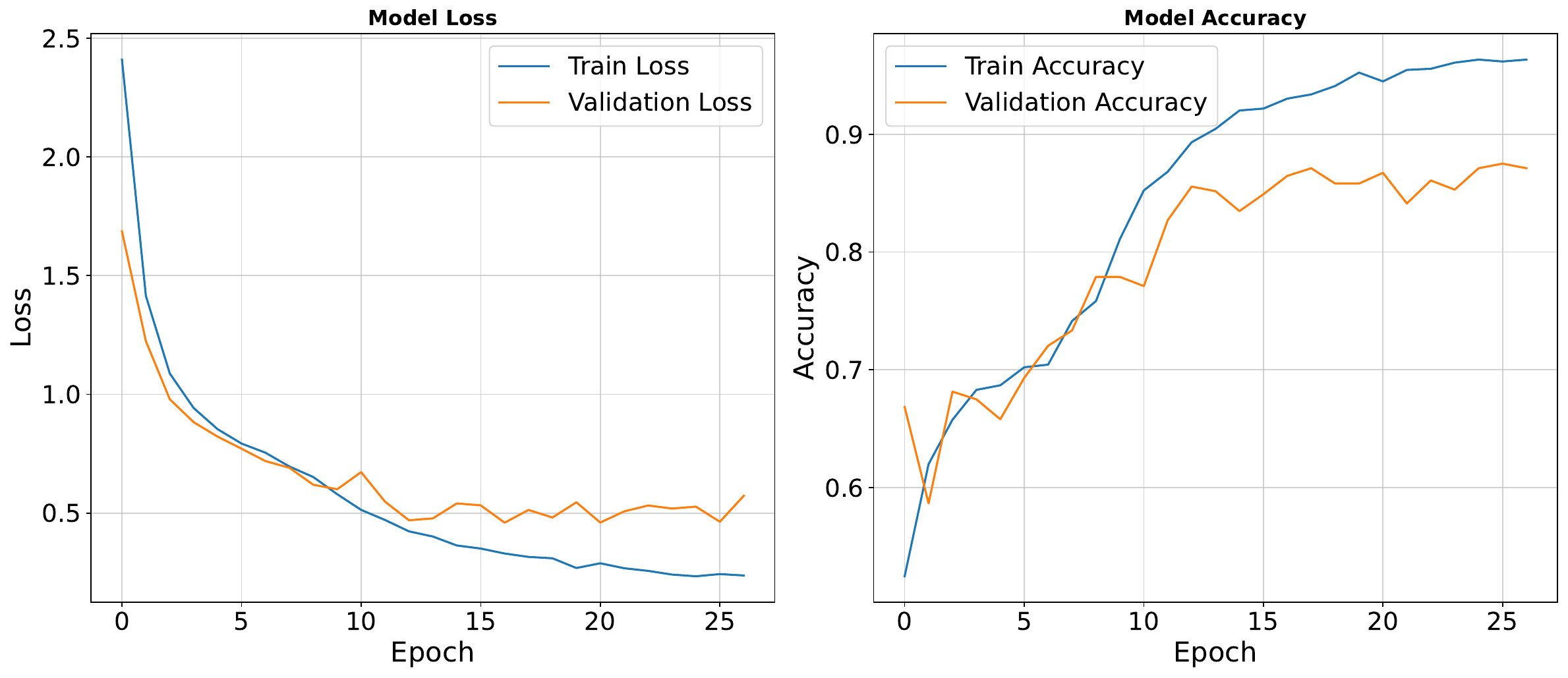}
  \caption{Evolution of training and validation loss and accuracy over epochs for the 7$\times$7 FFN matrix classifier trained \emph{only} on the filtered and balanced subdomain \(\Omega_0 = \{A : |c_0/c_1| < 0.08\}\).}
  \label{fig:perf-filtered}
\end{figure}

On this filtered and balanced subdomain \(\Omega_0\), we trained the same comprehensive hybrid FFN architecture used previously (73 features: 24 selected derived features DF1-DF5, DF7, plus 49 raw matrix entries), as detailed in Section~\ref{sec:derivedfeat}. Given the potentially more subtle distinctions required in this subdomain, a stronger L2 regularization (L2\_REG = 0.001) was applied to help prevent overfitting on this smaller, more focused dataset. The training evolution for this subdomain-specific model is shown in Figure~\ref{fig:perf-filtered}. The model trains effectively, reaching a high accuracy on this challenging subset.

%
%

Evaluating this subdomain-trained model on the (filtered and balanced) test portion of \(\Omega_0\) yielded the following confusion matrix:
\[
\begin{bmatrix}
\mathrm{TN} & \mathrm{FP}\\
\mathrm{FN} & \mathrm{TP}
\end{bmatrix}_{\text{FFN-}\Omega_0}
=
\begin{bmatrix}
1811 & 111 \\ 
55  & 1867  
\end{bmatrix}.
\]
This corresponds to an accuracy of approximately \( 95.8\% \) \textit{within this subdomain}. While direct comparison to overall accuracies is nuanced due to the filtering, the high accuracy here indicates the model can still learn effective distinctions even when the main separating feature \(|c_0/c_1|\) has been constrained.

\begin{figure}[htbp]
  \centering
  \includegraphics[width=0.9\textwidth]{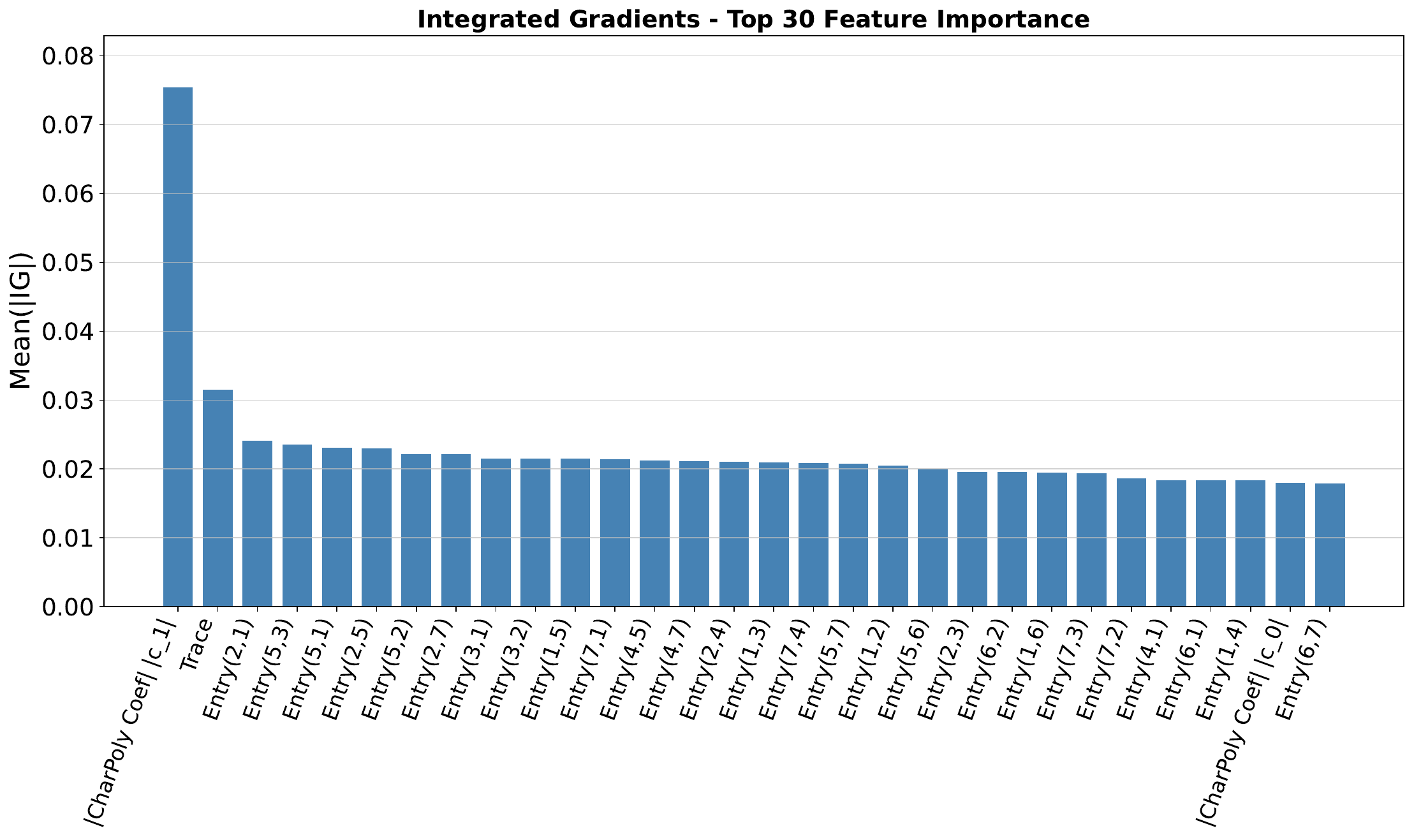}
  \caption{Feature importance for the FFN model trained on the filtered subdomain \(\Omega_0\), based on mean absolute Integrated Gradients. Note the shift in dominant features compared to Figure~\ref{fig:ig_summary}.}
  \label{fig:ig_summary_filtered}
\end{figure}

More notable is the change in feature salience when Integrated Gradients are applied to this subdomain-specific model. Figure~\ref{fig:ig_summary_filtered} presents the mean absolute IG scores for the top 30 features.

%
%
The salience map for the subdomain model (Figure~\ref{fig:ig_summary_filtered}) is radically different from that of the global model (Figure~\ref{fig:ig_summary}). Several key changes are observed:
\begin{itemize}
    \item The overwhelming dominance of \(|c_0|\) and \(|c_1|\) is significantly diminished. This is expected, as the dataset was filtered to have a narrow range of \(|c_0/c_1|\).
    \item Within this subdomain, \(|c_1|\) now appears as the most salient single feature,      more influential than \(|c_0|\) (which is now much lower in the ranking).
    \item The Trace (DF3) emerges as the second most influential feature.
    \item Raw matrix entries (\(A_{ij}\)) play a much more prominent role, with higher values of IG.
\end{itemize}
We infer that once the coarse separation provided by the \(|c_0/c_1|\) ratio is accounted for by focusing on \(\Omega_0\), the network learns to rely on a different set of features – primarily \(|c_1|\), the Trace, and specific raw matrix entries – to make finer distinctions.

A surrogate decision tree trained on the predictions of this subdomain FFN (using all 73 input features), 
selected features such as \(|c_0|\), \(|c_1|\), some matrix entries (e.g., Entry(4,2)), and the Fiedler value for its splits. Interestingly, despite the high IG salience of the Trace in the subdomain FFN, the decision tree surrogate did not prioritize it, showing the more challenging aspect of problem in the filtered domain.

This focused analysis on a subdomain demonstrates that once the main signal (\(|c_0/c_1|\)) is constrained, the network can learn more subtle patterns using other features. This approach of iterative refinement, where an initial coarse boundary defines a region for more detailed classification and explanation, opens a perspective for constructing hierarchical models or understanding complex decision boundaries in a piece-wise manner.
It also paves the way for future work exploring classification and explanation within truly different underlying matrix distributions, not just the uniform distribution explored here. For example, applying these techniques to specific subsets of structured matrices like M-matrices, or matrices arising from discretizations in finite element methods (which may have a very low probability of occurrence under a uniform \((-1,1)\) sampling but are of great practical interest), could yield tailored insights and potentially new, more specific criteria for monotonicity.

\section{Concluding Remarks} \label{sec:conclusion}

This work demonstrates how a pipeline combining deep learning with modern explainability methods can translate a high-performance "black-box" classifier into simple, interpretable mathematical rules. By applying this methodology to the long-standing algebraic question of what distinguishes a monotone matrix from the more prevalent non-monotone case, we uncovered a surprisingly simple structural principle. Starting from nothing more than i.i.d. sampling in $\mathcal{U}(-1,1)$, the network consistently focused on two global quantities—the magnitudes of the two lowest-order coefficients of the characteristic polynomial. Saliency maps, surrogate decision trees, symbolic regression, and linear SVMs all converged on the pair \((|c_0|,|c_1|)\), while every other spectral or entrywise feature played, at best, a supporting role. Compressed to its essence, the learned boundary is almost one–dimensional:
\[
    \left|\frac{c_{0}}{c_{1}}\right|\;\lesssim\;0.18,
    \qquad
    n=7,
\]
a statistical surrogate for the more cryptic requirement
\(A^{-1}\!\ge0\). Extreme-value theory supplied complementary evidence: within
our \(18\,000\) monotone samples the ratio behaves as a bounded random
variable, with a vanishing probability of exceeding the empirical maximum.

\medskip
\vspace{-0.3em}
\noindent\textbf{Summary of Findings.}
\begin{itemize}\setlength\itemsep{0.3em}
\item A high-accuracy classifier need not memorise \(\!49\) raw entries: two
      low-order coefficients suffice for \(\approx 95\%\) accuracy. Moreover, the models employing simpler feature sets demonstrated improved generalization compared to the full 73-feature hybrid model. The one-feature network, utilizing only the ratio \(|c_0/c_1|\) and achieving over 93\% accuracy (Figure~\ref{fig:perf-ratio}), showed exceptional generalization: its validation accuracy often remained almost always above the training accuracy, with validation loss  below training loss, patterns absent from the 73--feature curves, strongly indicating a robust signal capture. The two-feature model (\(|c_0|, |c_1|\)), while exhibiting some gap between training and validation metrics (Figure~\ref{fig:perf-2feat}), was also  significantly less prone to overfitting, reinforcing these simpler representations effectively discern the signal from the noise.

\item The trace of \(A^{-1}\) emerges, somewhat serendipitously, as a practical  diagnostic: large traces (small \(|c_0/c_1|\)) are a statistical signature
      of monotonicity in the uniform sampling space.
\item Once the coarse separation is enforced (via \(|c_0/c_1|<0.08\)), finer
      distinctions within this filtered subdomain hinge on \(|c_1|\), \(\mathrm{tr}(A)\), and now more contributing matrix raw  entries. This suggests that  deep within this low-ratio regime, monotonicity reflects more subtle  structural features.
\end{itemize}

\medskip
\vspace{-0.4em}
\noindent\textbf{Outlook.}
The present study is agnostic about matrix origin; every entry is equally
likely.  In real-world applications, however,  matrices are rarely uniform. However, a clean, controlled baseline was a necessary starting point: it allow us benchmark architectures, attribution methods, and statistical tests before applying them to the narrower (and often highly imbalanced) distributions that practitioners care about. 
Three directions merit further investigation:

\begin{enumerate}\setlength\itemsep{0.3em}
\item \emph{Structured subclasses.}  
      Develop targeted datasets for matrix types such as Z-matrices, M-matrices, 
      and Stieltjes matrices. Also consider stiffness matrices derived from specific 
      finite-element meshes. Do the same minimal features govern classification, 
      or does the inherent structure introduce new invariants?
      
\item \emph{Alternative priors.}  
      Replace the uniform cube with distributions reflecting physical or
      geometric constraints (e.g.\  heavy-tailed priors, sign-pattern
      priors).  How does the empirical threshold \(|c_0/c_1|\) shift under such distributions?
\item \emph{Beyond necessity.}  
      The ratio bound is a robust necessary condition for \(n=7\).  Can
      similar EVT-backed arguments turn other low-degree symmetric functions of
      eigenvalues into practical \emph{sufficient} tests, at least for   well-behaved subclasses?
\end{enumerate}

A dataset that purposefully \emph{over-represents} rare monotone
configurations—such as sparse M-matrices with controlled off-diagonal decay—would enable the same interpretability framework to isolate features that characterize specific families. If current trends persist, we may eventually identify a small set of low-dimensional signatures, each characterizing a distinct class of monotone matrices.

\section*{Acknowledgments}
To expedite text production, AI tools were used for editing and polishing the authors written work, focusing on spelling, grammar, and general style, and were subsequently meticulously reviewed, revised, and modified by the authors.

\paragraph{Data and code availability}
The HDF5 data sets and training scripts supporting this study are available from the corresponding author upon reasonable request.


\end{document}